\pdfoutput=1
\documentclass[a4paper,fleqn]{cas-dc}
\usepackage[round]{natbib}

\usepackage{multirow}
\usepackage{caption}

\usepackage[utf8]{inputenc}
\usepackage[T1]{fontenc}
\usepackage{booktabs}
\usepackage{array, caption, threeparttable}
\usepackage[font=small,labelfont=bf,labelsep=none]{caption}

\usepackage[section]{placeins}
\def\tsc#1{\csdef{#1}{\textsc{\lowercase{#1}}\xspace}}
\tsc{WGM}
\tsc{QE}

\begin{document}
	\let\WriteBookmarks\relax
	\def\floatpagepagefraction{1}
	\def\textpagefraction{.001}
	\let\printorcid\relax 
	
	\shorttitle{First Mapping the Canopy Height of Primeval Forests in The Tallest Tree Area of Asia} 
	
	\shortauthors{Guangpeng Fan et al.}
	
	\title[mode = title]{First Mapping the Canopy Height of Primeval Forests in the Tallest Tree Area of Asia}  
	
	
	\tnotetext[1]{This work was supported by Tibet Autonomous Region Science and Technology Plan Project (XZ202301YD0043C); Key Project of National Natural Science Foundation of China (4233000283); National Postdoctoral Innovative Talent Support Program (BX20220038).}

	\author[1,2]{Guangpeng Fan}
	\ead{fgp1994@bjfu.edu.cn} 
	\author[2]{Fei Yan}
	\fnmark[1] 
	\author[1]{Xiangquan Zeng}
	\author[1]{Qingtao Xu}
	\author[1]{Ruoyoulan Wang}
	\author[1]{Binghong Zhang}
	\author[1]{Jialing Zhou}
	\author[3]{Liangliang Nan}
	\author[4]{Jinhu Wang}
	\author[5]{Zhiwei Zhang}
	\author[2]{Jia Wang}
	\ead{wangjia2009@bjfu.edu.cn}
	\cormark[1]

	\address[1]{School of Information Science and Technology, Beijing Forestry University, Beijing 100083, China}
	\address[2]{Beijing Key Laboratory of Precision Forestry, Beijing Forestry University, Beijing 100083, China}
	\address[3]{3D Geoinformation Research Group, Faculty of Architecture and the Built Environment, Delft University of Technology, 2628 BL Delft, The Netherlands}
	\address[4]{Institute for Biodiversity and Ecosystem Dynamics (IBED), University of Amsterdam, Amsterdam, The Netherlands}
	\address[5]{Key Laboratory of Forest Ecology in Tibet Plateau, College of Resources and Environmental Sciences, Tibet Agricultural \& Animal Husbandry University, Nyingchi, 860000, China}
	
	\cortext[1]{Corresponding author} 
	\fntext[1]{Co-first author} 
	\begin{abstract}
		We have developed the world's first canopy height map for the primeval forest located within the distribution area of world-level giant trees, where the tallest tree in Asia (102.3m) was recently discovered. This mapping is crucial for identifying more individual and community world-level giant trees, as well as for analyzing and quantifying the effectiveness of biodiversity conservation measures in the Yarlung Tsangpo Grand Canyon (YTGC) National Nature Reserve under conditions of global warming. We proposed a method to map the canopy height of the primeval forest using deep learning base on the fusion of spaceborne LiDAR and satellite imagery (GEDI, ICESat-2, and Sentinel-2). We have developed a customized depthwise separable convolutional (DSC) neural network—PRFXception, which incorporates pyramid receptive fields. PRFXception, tailored specifically for mapping the canopy height of the primeval forest, efficiently integrates multi-size receptive field features to infer the canopy height at the footprint level of GEDI and ICESat-2 from Sentinel-2 optical imagery with a 10-meter spatial resolution. To validate the our proposed approach, we have conducted a field survey of 227 permanent plots in the giant tree distribution area using a stratified sampling method and measured several giant trees, including "Asia's tallest tree" and their communities using UAV-LS, and the predicted canopy height of the primeval forest was compared with ICESat-2 and GEDI validation data (RMSE=7.56m, MAE=6.07m, ME=-0.98m, $R^2$=0.58m), UAV-LS point clouds (RMSE=5.75m, MAE=3.72m, ME=0.82m, $R^2$=0.65m), and ground survey data (RMSE=6.75m, MAE=5.56m, ME=2.14m, $R^2$=0.60m). We mapped the potential distribution of world-level giant trees and discovered two previously undetected giant tree communities with an 89\% probability of having trees 80-100m tall, potentially taller than Asia's tallest tree. The multi-source Earth observation data-driven PRFXception deep learning integrated framework we propose is expected to achieve operational forward-looking monitoring of the height and dynamics of primeval forests worldwide. Combined with UAV-LS and field surveys, it provides a promising method and tool for discovering individual and community world-level giant trees. This paper provides scientific evidence confirming southeastern Tibet—northwestern Yunnan as the fourth global distribution center of world-level giant trees, which is crucial for supporting global climate and sustainable development initiatives and promoting the inclusion of the YTGC giant tree distribution area within the scope of China's national park conservation.
	\end{abstract}
	
	
	\begin{highlights}
	\item First mapping the primeval forest canopy height of the tallest tree growing in Asia
	\item Deep learning driven by multisource Earth observation to monitor the giant trees area
	\item Customized pyramid receptive field depth separable CNN, with an RMSE of 5.75m
	\item Found the world-level giant tree communities, could be taller than 102.3 m
	\item Provided a promising tool for discovering individual giant trees and communities
	\end{highlights}
	
	
	\maketitle
	
\section{Introduction}
\label{sec1}
\subsection{Discovery and survey of the fourth world giant tree area}
The southeast region of Tibet, China, has unique geographical and environmental characteristics in the world. It is the location of the longest and deepest canyon in the world - the Yarlung Tsangpo Grand Canyon (YTGC), which concentrates more than half of the biological species on Earth, and is one of the regions with the richest mountain forest ecosystem and biodiversity resources in the world. Intact primary forests are a symbol of the health and integrity of ecosystem productivity, an important indicator of the well-being of the natural environment, and a habitat for the tallest trees in the region \citep{ren2024conserving}. The tall giant tree is an important indicator of whether the natural environment of a region is good or not, and it is also a spiritual totem of the whole region. As a small ecosystem, the giant tree itself provides an excellent growth environment for arboreal organisms, and the active birds, fungi, insects and other organisms in the canopy are very rich. YTGC area has the world's least human disturbance of the original forest, is the largest intact original forest in China, there are tropical and temperate rain forests. In October 2021, a 72-meter balding fir was found in Gaoligong Mountain in northwestern Yunnan, China, becoming the tallest tree in China at that time. In May 2022, a 76.8 m Bhutanese pine and an 83.4 m yellow fir were discovered in Medog and Zayü County in southeast Tibet, setting a new record for tallest trees in China. In May 2023, a 102.3m Tibetan cypress was found in Bomi County, southeast Tibet, which was the first time that a 100-meter giant tree was found in Eurasia, becoming the tallest tree in Asia, second only to the American coast red shirt with a height of 116.1m, setting a new world record and becoming the second tallest tree in the world. The unique geographical environment of Southeast Tibet-Northwest Yunnan has given birth to a new world-level giant tree distribution center. At present, mainly in southeast Tibet, Bomi giant tree group, Medog giant tree group, Zayü giant tree group and Gongshan giant tree group in northwest Yunnan constitute the fourth world giant tree distribution area.
\par The fourth world-level giant tree distribution area is located in the eastern section of the watershed mountains on the Qinghai-Tibet Plateau, which is an important channel for the warm and humid water vapor of the South Asian monsoon to the north. The tallest trees in China and even Asia grow under superior hydrothermal conditions. Previously, only three places in the world have been recognized as capable of growing a variety of trees taller than 80m, including the Coast Mountains in North America stretching from northwestern California to the border between Washington state and British Columbia, southeastern Australia, and Southeast Asian rainforests. At present, more than 260 of 3 species of 80m-tall giant trees have been found in southeast Tibet, and 25 of them are more than 90m-tall. These giant trees represent the undamaged integrity and authenticity of the primary forest ecosystem in southeast Tibet. In the preliminary work of this paper, we investigated the fourth world giant tree distribution area mainly in southeast Tibet, and measured the height of several world giant tree individuals and communities represented by "the tallest tree in Asia" with UAV and ground LiDAR technology, and found that these giant trees contain a large amount of carbon. Proving that these giant trees can serve as a proxy for the high productivity of the YTGC region. 46 species of higher plants were found on the tallest trees in Asia, including 9 species of moss, 1 species of lycopods, 7 species of ferns and 29 species of angiosperms. Among the epiphytes, there are 8 kinds of Orchidaceae, including 2 new taxa, and 2 kinds of succulent plants from the Crassulaceae family. Under the vertical span of 100 meters, the arboreal plants realize a great change from drought to humid, from sun to shade tolerance, and there are huge differences in plant groups. The southeast region of Tibet is full of various "tree Kings", not only the world's tallest living tree, but also the king of rhododendron, the largest ormosia hosieiand other special trees in different categories, as well as giant plants such as Medog begonia. These amazing "discoveries" continue to emerge, suggesting the possible existence of other 'tallest' tree kings of different categories. In The 1990s, National Geographic magazine sponsored an expedition by Ian Baker, Ken Storm, and Hamid Sardar, who published their travel book The Heart of the World: A Journey to Tibet's Lost Paradise offers a memorable description of what may be the world's tallest tree along the YTGC. Although the measurement of giant trees has reached the 100-meter level in the fourth world distribution area, the survey scope only covers a small area, there are still a large number of unexplored areas, and it is very likely that there are giant trees higher than the tallest trees in Asia.
\par It is of great scientific value and significance to map the canopy height of the primary forest in the fourth world level giant tree distribution area for the discovery of individual and community of world level giant trees and the conservation of biodiversity. Forest canopy height is listed as a high priority biodiversity variable for spatial observations and is a strong predictor of species richness on a regional to global scale. Canopy height, as a fundamental variable characterizing forest structure, is known to be associated with important biophysical parameters such as biodiversity, primary productivity, and above-ground biomass. Canopy height models (CHMs) can characterize habitat heterogeneity as a key structural feature that, together with composition and function, helps monitor ecosystem response to climate and land use change, as well as recovery. Studies have shown that forests buffer the microclimate temperature under the canopy, and higher forest canopies have a stronger inhibition effect on extreme microclimate, but whether this relationship is also applicable in the range of primary forests, where the world-level giant trees that are the tallest trees in Asia are born, needs further targeted research. Therefore, uniform, high-resolution CHM has the potential to advance modeling of climate impacts on ecosystems in southeast Tibet, and may assist the management of the Brahmaputra National Nature Reserve and strengthen microclimate buffers to protect biodiversity in a warming climate \citep{asner2014mapping}. At present, remote sensing monitoring and field verification of primary forests in the newly discovered fourth world level giant tree distribution area are very weak and scarce. The previously published map of forest canopy height in China with a spatial resolution of 30m shows that the value of this area is greater than 30m, but there is saturation effect or insensitivity for higher canopy distribution \citep{schwartz2024high}. The large-area, high-precision and all-round mapping of forest canopy height in the distribution area of world-level giant trees is of great significance for discovering higher "tree Kings", biodiversity protection and monitoring, and the impact of climate change on primary forests with unique geographical and ecological environmental characteristics of the world.
\subsection{LiDAR remote sensing of forest canopy height}
\par Earth observations and satellite remote sensing provide data for monitoring the quantity and quality of forests at regional and global scales and play a key role in mapping forest canopy height \citep{pang2022retrieval}. As an active remote sensing technology, Light Detection and Ranging (LiDAR) provides a method different from the field measurement of tree height, and obtains accurate three-dimensional forest structure information by emitting laser pulses. Traditionally measured in the field using altimeter, Brules altimeter and other tools to measure tree height, recently more widely used near ground LiDAR (UAV, airborne, backpack, handheld, terrestrial). Both methods can accurately measure tree height from individual trees to stand scales, but both are expensive and impractical over large geographic ranges. Space-borne LiDAR provides a way to measure forest canopy height at regional or global scales, such as the Ice, Cloud and IandElevation Satellite (ICESat), GLAS, ICESat-2 ATLAS and GEDI. In 2003, NASA launched the world’s first Earth observation laser altimeter satellite in the Earth Observing System EOS (Earth Observing System) \citep{he2023icesat}. GLAS plays an irreplaceable role in the remote sensing application of polar ice and snow environment, notably in the measurement of ice sheet elevation and sea ice thickness. L.I. Duncanson developed a method to consistently simulate forest height directly from GLAS waveform measurements and determined the utility of topographic relief affecting GLAS waveform indicators \citep{duncanson2010estimating}. Yanqiu Xing proposed a method to improve the estimation accuracy of the maximum canopy height \citep{xing2010improved}. GLAS waveform on inclined terrain was used to reduce the mixed effect of inclined terrain and rough terrain, and the result greatly improved the accurate estimation of the maximum canopy height on inclined terrain. ICESat-2, launched in 2018, carried the ATLAS load to continue the unfinished tasks of ICESat, which was mainly used to measure the surface elevation, including the elevation of ice sheets, forests and oceans, and reveal the status quo and change rules of vegetation biomass in large areas. ICESat-2 provides a comprehensive estimate of global topography and canopy height and is a new opportunity to directly measure the height and distribution of boreal forests. Xiaoxiao Zhu showed that all ICESat-2 data were suitable for extracting ground elevation and forest height \citep{zhu2020performance}. The performance of ICESat-2's weak beam was worse than that of the strong beam, and the performance of ICESat-2's daytime data was worse than that of the night data, and the strong beam at night was more suitable for estimating forest canopy height. Neuenschwander analyzed the accuracy of the preliminary ATL08 data product for the flight mission by using ICESat-2's plane surface over Finland's vegetation cover area, and showed that there was a 5-meter deviation (horizontal accuracy) in geolocation that fully met the mission requirement of 6.5 meters \citep{neuenschwander2019canopy}. The vertical root-mean-square error of canopy height inversion is 3.2m. GEDI's measurements of forest canopy height, canopy vertical structure, and surface elevation can greatly improve the ability to characterize biodiversity, carbon and water cycle processes, and nature reserves. Li accurately estimated the canopy height of sparse grassland areas between 3 and 15 meters by comparing the GEDI data in orbit with the simulated GEDI RH98 index derived from the reference ALS dataset \citep{li2023first}. Quiros compared GEDI data in southwest Spain with airborne LiDAR data, and analyzed the consistency of airborne LiDAR data, GEDI footprint, and canopy height \citep{quiros2021gedi}. Globally, the RR100 achieved an RMSE value of 3.56m, which is in better agreement with airborne LiDAR data.
\par However, there are limitations in the mapping of forest canopy height with spaceborne LiDAR. The discrete ground sampling footprint represents a limited point sample of the land surface, most of which is not observed and covered. For example, ICESat GLAS's coverage ranges are about 170 m from each other along a single track and tens of kilometers across multiple tracks \citep{li2024performance,silva2021fusing}. The ICESat-2 ATLAS carries two lasers, usually only one of which is in operation, and can obtain overlapping spots about 0.7 m apart and 17 m in diameter along the orbit. The 6 laser beams emitted by ATLAS are arranged in three parallel groups along the direction of the orbit, with a cross-rail distance of about 3.3km between each group and a cross-rail distance of about 90m within the group \citep{feng2023systematic}. The GEDI instrument consists of three lasers that produce a total of eight beam ground cross sections, including footprint samples of about 25 m spaced 60 m along the orbit, beam cross sections spaced about 600 m across the Earth's surface along the direction of the cross orbit, and a cross-orbital width of about 4.2 km, with limited geographic range, spatial and temporal resolution. GEDI is expected to cover up to 4 percent of the land surface, and the samples collected are sparsely covered on the Earth's surface, which limits the resolution of the grid mission product to 1-kilometer units \citep{nelson2017lidar}. One of the scientific objectives of both ICESat-2/ATLAS and GEDI includes forest height estimates, and Liu evaluated forest heights obtained from GEDI and ICEST-2 data in forests in the northern, central, and western United States \citep{liu2021performance}. The study showed that the combination of ICESat-2 and GEDI provided the lowest forest height, with RMSE of 5.02 and 3.56 m, respectively. GEDI and ICESat-2 ATLAS have a much higher footprint density than ICESat GLAS, but they still cannot directly provide a comprehensive (wall-to-wall) view of forest canopy height \citep{brandt2023wall}. The data collection time of GEDI and ICESat-2/ATLAS is roughly the same \citep{zhu2022consistency}. The data fusion of GEDI and ICESat-2 / ATLAS can increase the density of sample points at forest height and realize the geographical space complementarity.
\subsection{Optical remote sensing of forest canopy height}
\par To overcome the space discontinuity problem of spaceborne LiDAR, sensor fusion between spaceborne LiDAR and multispectral optical imaging has the potential to overcome the limitations of separate data sources. The researchers attempted to fuse spaceborne LiDAR with optical images to create a continuous map of forest canopy heights. Landsat, ZY-3, MODIS, Sentinel-2 and other satellite missions can meet a wider range of Earth observation needs, providing freely accessible optical images \citep{spawn2020harmonized}. These satellite images are not tailored to the structure of the vegetation, but provide longer-term global coverage at high spatial and temporal resolution. Estimating forest features such as canopy height or biomass from optical images is a challenging task because the physical relationship between spectral features and vertical forest structure is complex and not well understood. Optical images may have a strong saturation effect in forests with large canopy density, and the product of canopy height estimated by optical images is often full of uncertainty \citep{tang2019characterizing}. Existing standard tools for forest canopy height measurements tend to underestimate, as the height saturation is estimated to be around 25 to 30 meters. This is a rather severe limitation, especially in high canopy primary forest areas, and worsens downstream carbon stock estimates because giant trees have particularly high biomass. Potapov combined optical image (Landsat) multi-surface reflection data with LiDAR (GEDI) canopy height indicators to create a global map of forest canopy height in 2019 with a resolution of 30 meters \citep{potapov2021mapping}. Another limitation of previous large-scale CHM projects was the reliance on local calibration, which hindered their use where no reference data was available. Lin proposed a method to map forest canopy height at regional and global scales \citep{lin2020estimates}. By combining ICESat-2 atlas data with ZY-3 stereoscopic images, multiple data sources can complement each other and jointly map regional forest canopy height, providing topographic information to cover the whole world at a low cost. Malambo developed a gradient-enhanced regression model that correlated canopy height with auxiliary data values, and predicted the canopy height of unobserved locations with a spatial resolution of 30m \citep{malambo2022landsat}. It is demonstrated that the fusion of ICESat-2 orbital canopy height estimation with Landsat and LANDFIRE data can generate spatially complete canopy height data at the regional level in the United States.
\par The researchers tried to fuse different remote sensing data sources into a dense canopy height map, mainly considering the availability of canopy heights in some places, and the dense coverage of optical satellite sensing over large areas \citep{waldeland2022forest}. Technically, this is equivalent to regression of crown height from monocular (multispectral) images using known tree heights as reference data, and the height of the "ground reference" tree can be derived from LiDAR, or it can be derived by gathering enough field observations. Hudak used airborne LiDAR as ground truth to demonstrate the Kriging calculation of vegetation height from the original per-pixel Landsat spectra \citep{hudak2002integration}. The processing by Ota is technically more similar to a model using a medium-resolution sensor, regression canopy height from an "interference indicator", i.e., pixel statistics of an annual time series calculated on a Landsat image with a brightness/greenness/humidity conversion \citep{ota2014estimation}. Tyukavina regression canopy height from raw Landsat data and GLAS measurements to quantify forest carbon loss over time to a MAE of 5.9 m \citep{tyukavina2015aboveground}. A tree-based regression algorithm in subsequent work has been applied to the same data sources, using the multi-temporal features of each Landsat channel \citep{hansen2016humid} for a 1° wide north-south cross-section in tropical Africa, with a MAE of 2.5 m (mean tree height \textless 10 m). In forest areas with crown height, the height is systematically underestimated, and the MAE of trees with 5m height is $\approx$ 25 m, and that of trees higher than 13 m is \textgreater 30 m. As a result, the generated map is saturated above 25 m. All of this work is based only on (in some cases multi-temporal) spectral data from a single pixel location, although plane textures may have the potential to serve as further proxy signals to reveal forest vegetation structure information, especially at high resolution \citep{hoang2021mapping}. For a tree with a height of 30 m, the root-mean-square error is $\approx$ 2 m, while if there is a tree with a height of 40 m, the accuracy drops to $\approx$ 4 m. Previous studies have shown that the inversion of tall trees is mainly supported by optical Landsat images. Previously, satellite-based forest canopy height Earth observation data were limited to separate data sources such as GLAS and ATLAS, and more recently, to GEDI data sources \citep{valbuena2020standardizing}. Sentinel-2 has high-resolution observation capabilities with a maximum resolution of 10 meters and has technical advantages in terms of spatial, temporal and spectral resolution when operating with a shorter access period of 5 days. But relatively few studies have used Sentinel-2 to map forest structural features to date, possibly because the time series is still short. According to this statistical survey, there are probably only a few studies that have tested Sentinel-2 \citep{astola2019comparison} for height prediction in boreal forests, training two layers of perceptrons on a small number of tiles (\textless 200). Sentinel-2 was found to have a slightly lower prediction error than Landsat, and one does not pay a performance price for obtaining higher spatial resolution. Several studies have used regression trees to evaluate the retrieval of other forest biophysical parameters, including growth stock of Mediterranean forests \citep{chrysafis2017assessing}, and generalized linear models \citep{korhonen2017comparison} to extract canopy coverage and leaf area index (LAI) of boreal forests. Interestingly, the study reports that the prediction performance directly using Sentinel-2 bands is at least as good or better than the vegetation index derived from the data.
\subsection{Challenges of remote sensing of canopy height in primary forest}
\par At present, there are relatively few attempts to map the canopy height of primary forest in large areas by combining GEDI, ICESat-2 and Sentinel-2. The challenges of mapping the canopy height of primary forest with satellite-borne LiDAR fusion multi-spectral images mainly include satellite-borne LiDAR waveform processing of complex terrain of primary forest, band index performance of multi-spectral images, coupling relationship and processing of different data sources, accuracy and performance of regression methods \citep{dixon2023satellite}. The primary forest is a multilayer natural forest, which usually has a distinct vertical structure, including above, under and tussock, and shows significant differences in structure, phenology and photosynthetic capacity. It is still challenging for spaceborne LiDAR to extract upper and understory waveforms of forests with different canopy coverage and terrain conditions. Waveform processing must determine features such as the position of the ground and crown tops to derive crown height and other structural variables such as crown cover and vertical profile. The effects of known (e.g., pulse shape, digitizer noise) and unknown (e.g., slope, multiple scattering) characteristics are often difficult to model explicitly when separating the canopy and ground signals and determining ranging points along the waveform. For example, under dense canopy conditions, ground echoes can be quite weak. It is difficult to detect such a weak signal under high background noise. The detected tip of the tree may have very little leaf area, so the return is weak, which will lead to an error in height recovery. Nico Lang proposes a data-driven approach based on state-of-the-art probabilistic deep neural networks that can estimate global canopy height as well as well-calibrated uncertainties in GEDI waveforms in geolocation (class 1B/L1B), improving retrieval performance \citep{lang2022global}. Yun extracted upper and understory waveforms using GEDI L1B data, taking into account multiple peaks of waveforms, canopy coverage, and topographic slope, and investigated the effects of these factors on forest stratification \citep{yun2023stratifying}. A canopy height map with a ground resolution of 10 m is presented using multispectral data regression, at which the spectral features of a single pixel are no longer sufficient to predict tree height. Physical phenomena behind tree height predictions, such as shadows, roughness, and species distribution, cause reflection patterns that span multiple pixels. However, it is not obvious how to encode the resulting image textures into predictive feature descriptors that support regression.
\par The regression method is very important for mapping the forest structure characteristics on a pixel-by-pixel basis. Previous studies have used simple linear regression (SLR), multilinear regression (MLR), regression trees, support vector regression (SVR), and random forests (RF) for regional scale forest structure estimation. Based on research data sets, it is found that RF or SVR perform better in interpreting nonlinear relationships between forest structures, such as mean canopy height \citep{simard2011mapping} or stock volume \citep{hu2020estimating} and remote sensing multispectral reflectance \citep{tolan2024very}. For example, multispectral features from MODIS (Moderate Resolution Imaging Spectroradiometer) and climate variables are used to map global forest canopy heights. Multi-spectral features and texture indices calculated by Sentinel-2 and PALSAR-2 were used to construct an RF model for mapping forest stock at regional scale. Sothe proposes a simple way to integrate ICESat-2 and GEDI, combining synthetic aperture radar and optical data from Sentinel 1, Sentinel 2, and the United States Commission on the Law of the Sea PALSAR-2 to generate a map of forest height distribution in Canada \citep{sothe2022spatially}. The study compared data from two new spaceborne LiDAR sensors, ICESat-2 and GEDI, to estimate forest canopy heights in Canada. LiDAR data are correlated with Sentinel-1 and Sentinel-2 and ALO-2/PALSAR-2 to produce a continuous canopy height map with a resolution of 250 meters. For RMSE, GEDI had an error of 4.2 meters, 1 meter lower than the observed ICESat-2. Wang's fusion of spaceborne LiDAR and optical images improves the accuracy of canopy height estimation, and effectively improves the problems such as the limitations of complex terrain, the impact of dense vegetation, and the imbalance of spatial accuracy in forest canopy height mapping. Many studies have shown that the fusion of satellite-borne LiDAR and remote sensing image data has great potential in forest canopy height mapping \citep{ene2012assessing}. Canopy height in the coverage area of spaceborne LiDAR was treated as a ground reference, and a regression model was established using optical images and environmental variables as predictors. For example, Lefsky used Cubist method as a regression model and used ICESat GLAS data and MODIS images to predict the global forest canopy height distribution at forest patch scale \citep{lefsky2010global}. Simard used random forests as a regression model, and used ICESat GLAS data, tree cover and environmental variables derived from MODIS to generate a global forest canopy height product with a resolution of 1 km. Potapov used the bagged regression tree set method for regression and mapped the global distribution of forest canopy height at 30 m resolution using GEDI, Landsat8 OLI images and phenological indicators. In essence, these studies are similar to methods used to estimate forest canopy height using field measurements and optical images, which rely on optical images to achieve continuous mapping objectives \citep{hao2021automated}. Although a large number of spaceborne LiDAR footprints can improve the estimation accuracy of the regression model, the generated forest canopy height product still has a saturation effect. In the process of model training, GEDI data will have data bias, and some regional models will have higher results. The research uses a random subset of training data for calibration, creates a local calibration training model, and also needs to maintain the consistency of the global map. Potapov predicted forest height values based on spatiotemporal multispectral Landsat data based on a machine learning algorithm per pixel (regression tree). The regression tree model was calibrated and applied in a "moving window" mode to each individual Landsat GLAD ARD block (1 × 1°). This strategy does not make full use of spaceborne LiDAR data, especially the more recent GEDI and ICESat-2. During the regression process, the precise forest canopy height information in the satellite-borne LiDAR coverage area may become blurred.
\subsection{Depth regression model in remote sensing}
\par In recent years, machine learning has been increasingly used to model different types of data (individually or in combination) from LiDAR, field measurements, and optical images for monitoring forest structural features. Potapov used the relative height value of GEDI percentile and Landsat multi-temporal index to calibrate the regression tree model based on the machine learning algorithm per pixel (regression tree), and predicted the forest height value based on spatiotemporal multi-spectral Landsat data. Wang Li used deep learning model and random forest model to compare the performance of Sentinel and Landsat-8 satellites, and combined ICESat-2 and high-resolution sentinel satellite images with Landsat-8 images to draw the spatial pattern of forest canopy height in mountainous areas of China \citep{li2020high}. With the help of sentinel data auxiliary variables, DL model and RF model have achieved satisfactory results in ICESat-2 canopy height inversion. Previous studies have often used optical images to develop spatially continuous forest canopy height distributions, but optical images cannot take full advantage of the dense spaceborne LiDAR footprint, and may also be affected by the optical image saturation effect \citep{lopatin2016comparing}. Xiaoqiang Liu developed a new neural network guided Interpolation (NNGI) method to map forest canopy height by fusing GEDI, ICESat-2 maps, and Sentinel-2 images \citep{liu2022neural}. The fusion of GEDI and ICESat-2 data provides a more accurate value of forest canopy height, with an average forest canopy height of 15.90 meters in China and a standard deviation of 5.77 m. Recent advances in deep learning in computer vision and image analysis in the last two years have impressively demonstrated that deep convolutional neural networks (CNNS) can learn from raw images, with sufficient training data, multi-level feature coding tailored to a given prediction task.
\par Deep learning using convolutional neural networks has become the dominant technique for image analysis, including image-level classification, pixel-level semantic segmentation, and continuous variable regression. Kumar compared the performance of the DCNN model with multi-sensor fusion images and a single sensor dataset \citep{kumar2023comparative}. The results show that the DCNN model with fused data has high classification accuracy, reaching 99.8\% and 99.2\% on training and validation data sets, respectively. Deep separable convolutional neural network is based on convolutional neural network to optimize the model through more parametric efficient convolutional structure, so that it is more suitable for deployment in resource-limited environment \citep{mulverhill2022evaluating}. Deep neural networks have demonstrated their ability to model nonlinear relationships and have been successfully used to assist interpolation algorithms to predict air quality, which provides a potential solution for forest canopy height regression driven by multi-source remote sensing data to automatically adjust spatial, spectral and environmental differences. There are relatively few studies on pixel-by-pixel regression of forest canopy height in large geographical range using deep convolutional neural networks. End-to-end learning of rich contextual feature hierarchies is fundamental to success in image and raster data analysis, including object recognition, natural speech processing, and more. So far, most of the work where deep learning has been used for remote sensing has been for classification tasks such as land cover mapping. Although neural networks are a general-purpose machine learning technique that can solve classification and regression tasks using the same machines, relatively few studies have used them to retrieve continuous biophysical variables or metrics. Texture features are particularly important in primary forests or high-canopy forest areas. Nico Lang studied a deep convolutional neural network (CNN) to extract appropriate spectral and structural features from reflected images, and then returned the height of vegetation per pixel, extending the sensitivity of the regressor to a height of up to 55m \citep{lang2019country}. He proposes a new supervised probabilistic learning method based on depth separable convolutional neural networks (CNN) to interpret GEDI waveforms and regression global canopy top heights to avoid obvious modeling of unknown effects such as atmospheric noise.
\par Given the large amount of data collected by ICESat-2 and GEDI, uncertainties due to the lack of understanding of the physical mechanisms between spectral features and vertical forest structures are minimized by utilizing supervised machine learning, especially end-to-end deep learning \citep{narine2019synergy}. Although the retrieval of vegetation characteristics through deep learning has recently begun to be proven, the step of mapping the canopy height of the primary forest in the distribution area of the world-level giant tree that grows the tallest tree in Asia is full of unknowns and technical challenges in an area with unique geographical and ecological characteristics in the world \citep{urbazaev2022assessment}. In the development of neural networks, to improve the model's ability to understand complex data, introducing the concept of multi-Receptive Field became a key breakthrough. The concept of receptive fields is crucial to understanding how networks process and fuse information at different scales, especially in complex image processing tasks. The receptive field is a region of input data that neurons can "see". In the depth-separable Convolution (DSC) architecture of existing methods, the receptive field is single, which makes it difficult to extract multi-scale information, so that the classifier can take into account the learning of different fineness information. In remote sensing image processing, the neurons of a convolutional layer may perceive only a small part of the image. The size of the receptive field is determined by the size of the convolution kernel \citep{horache20213d}. A larger convolution kernel means a larger receptive field and can capture a wider range of spatial information. With the deepening of the network layer, the effective receptive field of each neuron also increases, and more layers of information can be integrated. Receptive field is one of the key concepts of convolutional neural networks, which directly affects network design and performance optimization \citep{ngo2023tropical}. The concept of multiple receptive fields in neural networks refers to the ability of the network to capture multi-scale features of input data through different sizes or different types of receptive fields. This design allows the network to understand both the local details of the input data and broader contextual information at the same time, improving the model's ability to understand and process complex data. This paper overcomes the challenge of selecting suitable convolution kernel size in convolution operation, combines the advantages of Inception and Xception, and increases the ability of extracting multi-scale features while reducing the number of parameters \citep{chollet2017xception,szegedy2015going}. Considering that stacking convolution layers of different sizes from "depth" is easier to overfit while consuming computing resources, a convolutional parallel structure with different receptive fields is designed in a customized deep separable convolutional neural network architecture. Running convolution operations with multiple sizes at the same level makes depth-separable convolutional neural networks "wider" rather than "deeper", avoiding randomness or uncertainty caused by blind application of existing network architectures in computer vision as much as possible.
\subsection{Our contribution and innovation}
\par To map the height of the primary forest canopy in the recently discovered fourth world-level giant tree distribution in high resolution, we propose a deep learning method called PRFXception, which is a custom-made depth-separable convolutional neural network that incorporates the pyramid receptive field mechanism. The proposed deep regression network is designed for the first time for the Earth observation mission data of spaceborne LiDAR (GEDI, ICAESAT-2) fusion of multi-spectral images (Sentinel-2). As a spaceborne LiDAR fusion optical image deep learning method described in this work, PRFXception uses publicly available spaceborne LiDAR and optical satellite images as inputs to map high-resolution primary forest canopy heights in southeast Tibet based on GEDI and ICESat-2 and Sentinel-2 data. From 2,832,791 data examples, our model learns to extract patterns and features that predict high-resolution vegetation structure from raw satellite images. By fusing ICESat-2 and GEDI altitude observations i.e., RH98, the relative height of 98\% energy return) with Sentinel-2 images, our approach enhances the temporal and spatial resolution of primary forest canopy heights with globally unique geographical features. And applied it to the fourth world giant tree range, which grows the tallest tree in Asia. We deployed the method to calculate canopy height with a 10-meter ground sampling distance (GSD) based on 2020 GEDI, ICESat-2, Sentinel-2 optical image data. The UAV-LS point clouds and ground survey data collected at the same time resolution were used to quantify the uncertainty of the model. Evaluation of forest canopy height products generated by PRFXception using spaceborne footprint data, unmanned aerial LiDAR data, and field sample survey data. We used stratified sampling method to investigate 227 permanent plots in southeast Tibet, and flew more than 10,000m\textsuperscript{2} of unmanned aerial LiDAR point clouds to verify the method proposed in this paper, and mapped the height of primary forest canopy in the fourth world-level giant tree distribution area with a resolution of 10m. We plan to make the map publicly available to support climate change, carbon and biodiversity conservation efforts in the region, and expect to make the map available to view interactively in a browser application soon.
\par To our knowledge, this paper is the first work dedicated to mapping the height of primary forest canopy in the newly discovered fourth world-level giant tree distribution area, and publishes the first global map of primary forest canopy height in the world-level giant tree distribution area. For the first time, we proposed a pyramid perception-field depth-separable convolutional neural network, and used it to drive deep learning modeling with multi-source observation mission data (GEDI, ICESat-2, Sentinel-2) to map the height of primary forest canopy in the pixel-level world-level giant tree distribution area. Our main contributions are :(1) For the first time, we mapped the height of primary forest canopy in the world's largest tree distribution area where Asia's tallest tree grows; (2) For the first time, deep learning modeling driven by GEDI, ICESat-2 and Sentinel-2 fusion was used to monitor global biodiversity hotspots in the Himalayas and verified with previously unpublished ground survey data of the same time resolution; (3) A depth-separable convolutional neural network (PRFXception) coupled with pyramid receptive field mechanism was specially customized to predict the forest canopy height with 10m resolution; (4) PRFXception was used for the first time to predict the height of the largest and best preserved primary forest canopy in China with globally unique geographical features, with an RMSE of 5.75m; (5) In our potential distribution map of giant trees, we found two previously undiscovered world level giant tree communities with an 89\% probability of being between 80 and 100m tall, possibly containing individuals taller than the tallest trees in Asia; (6) The proposed fusion of multi-source Earth observation mission data-driven deep learning modeling method is a promising tool for discovering world level giant tree individuals and communities, and (7) provides a strong scientific proof for confirming Southeast Tibet-Northwest Yunnan as the fourth world level giant tree distribution center. The YTGC National Nature Reserve will be included in China's national park protection.
\par The structure of this paper is as follows. Section 2 introduces the study area and data. Section 3 introduces the PRFXception architecture design, modeling process, experiment setup and model parameters of multi-source Earth observation mission data. Section 4 describes the results and discussion. Finally, the conclusions are presented in Section 5.
\section{Study area and data set}
\label{sec1}
\subsection{Study area}
\label{subsec1}
\begin{figure*}[tp]
	\centering
	\includegraphics[width=1\textwidth]{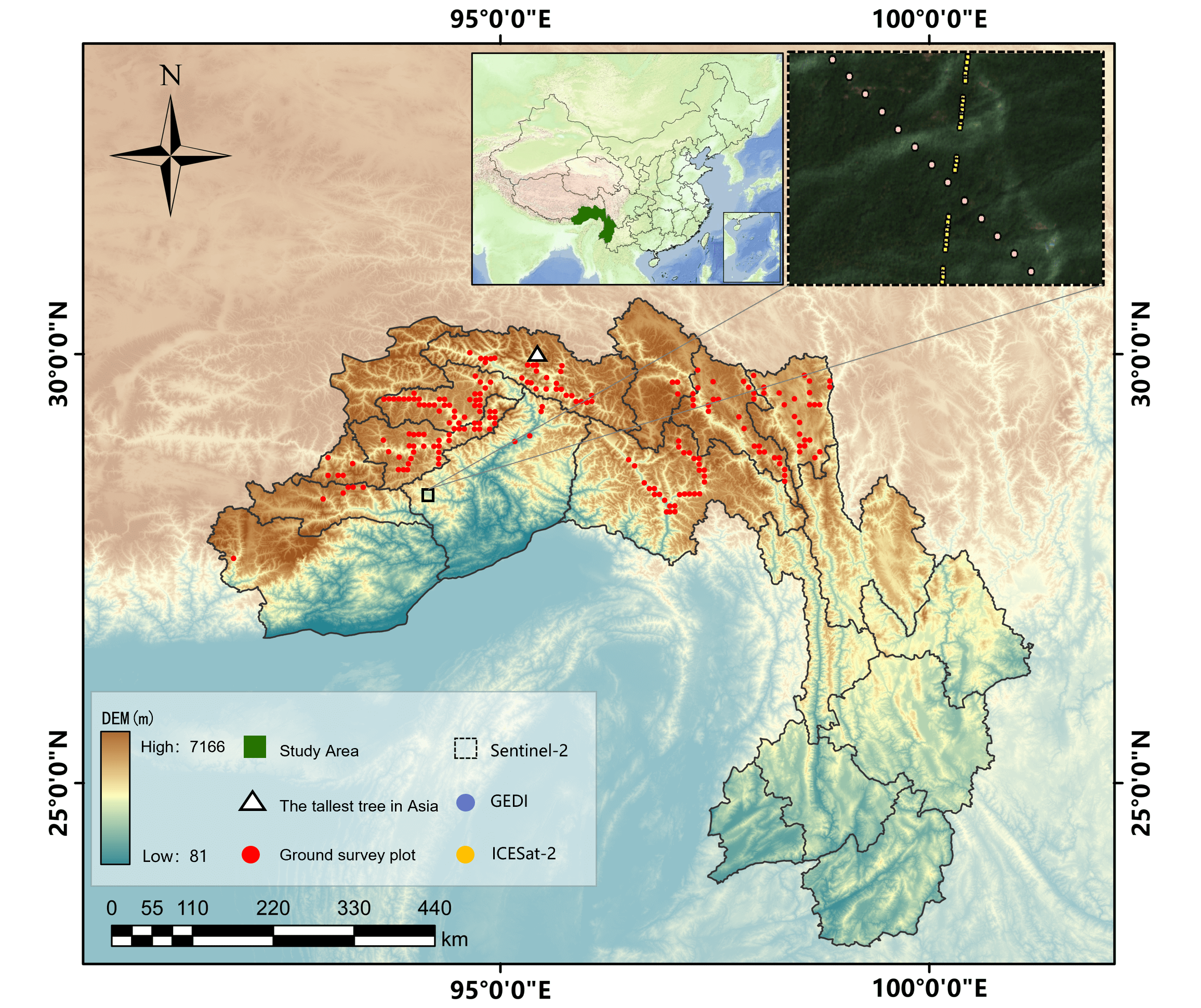}
	\caption{Study area map}
	\label{Fig. 1}
\end{figure*}

This paper compares the southeast region of Tibet (91°6 '50 "\textasciitilde 99°34' 22" E, 26°29 '52 "\textasciitilde 31°7' 33" N) with the northwest region of Yunnan Province (97° 03' 00" -100° 31' 00" E, 23° 53' 04"-29° 42' 00"N) consists of a fourth world-level giant tree center as a research area Fig. \ref{Fig. 1}. Southeast Tibet covers an area of more than 190,000 km\textsuperscript{2}, and there is the deepest and longest river canyon in the world - YTGC. From the highest peak of Namcha Barwa peak 7782m, to the south of the lower reaches of the Brahmaputra River plain area less than 100 m, the height drops of more than 7,000 meters. There are significant differences in vertical climate within the region, and the climate zone can be divided into tropical, subtropical, warm temperate zone, temperate zone, alpine tundra zone and ice and snow zone. The southeast is wet and rainy, while the northwest is cold and dry. Under the combined influence of altitude and climate factors, from southeast to northwest, the area with forest as the main land cover type transits to the area with grassland as the main land cover type. Under the influence of warm and humid air currents from the Bay of Bengal and the uplifting Tibetan Plateau, southeastern Tibet has abundant rainfall throughout the year, making it one of the wettest regions in the world. Abundant rainfall, suitable light, and good matching of water and heat conditions provide favorable conditions for breeding world-level giant trees. The YTGC area is the largest intact virgin forest in China. This unique area with both tropical rain forest and temperate rain forest is easier to grow world-level giant trees due to superior hydrothermal conditions, and even the tallest trees in the world may exist. The YTGC serves as a water vapor channel that allows the rainforest's supertall tree species to occupy the lowlands of southeastern Tibet, where the \emph{Altingia excelsa Noronha}
 is found, A supertall tropical tree such as \emph{Ailanthus integrifolia subsp.} \emph{calycina (Pierre) Noot}. In the high elevations of the region, there are \emph{Cupressus austrotibetica, Abies ernestii var. salouenensis, Pinus bhutanica Grierson, D.g.ong \& C.N.age, Taiwania cryptomerioides Hayata} and other supertall conifers. In 2023, a 102.3 m tall cypress tree was found in Tongmai Town, Bomi County in the region, the first time a 100-meter giant tree was found in Eurasia, becoming the tallest tree in Asia, setting a new world record and becoming the second tallest tree in the world by tree species. Therefore, southeast Tibet is a unique region in the world with both ultra-high broad-leaved trees (tropical) and ultra-high conifers (temperate rain forest). It is the region where the individual and community of giant trees constantly set new records in China and the world at present, and it is also the main research area monitored by Earth observation mission in this paper.
\par Northwest Yunnan covers an area of more than 140,000 km\textsuperscript{2}, ranging from 400 meters to 6,000 meters above sea level. Northwest Yunnan is located at the junction of the first and second steps of Chinese terrain. The territory is crossed by a number of rivers, such as the Nujiang River, Lancang River and Jinsha River, which are mainly influenced by the Hengduan Mountains. These rivers and mountains cut numerous high mountains and deep valleys, with large altitude drop and unique and changeable terrain, showing various features such as high mountains, plateaus and steep canyons. The northwest Yunnan region presents the rich diversity of plateau climate, mountain climate and monsoon climate. The temperature in the plateau region is moderate in summer and cold in winter, while the mountainous terrain leads to significant differences in the vertical direction of temperature and precipitation. Monsoon influences bring relatively high rainfall in summer and dry in winter. The northwest Yunnan region shows the characteristics of large temperature difference between day and night, which is particularly significant in the plateau region. Local differences in topography lead to local differences in climate. For example, the Nujiang Valley area is relatively warm and rainy, while the alpine area has low temperature and relatively little precipitation. The diverse terrain and climatic conditions together shape the rich ecological landscape in Northwest Yunnan. In areas such as the Nujiang Valley and Hengduan Mountains, there may be some giant and long-aged trees, may include \emph{Cunninghamia lanceolata (Lamb.) Hook}, \emph{Abies fabri (Mast.) Craib}, \emph{Metasequoia glyptostroboides Hu et Chen} and other trees. At the end of 2021, a 72-meter-high Taiwania cryptomerioides Hayata was discovered on Gaoligong Mountain in the area, which was recognized as the "tallest tree in China" at the time.
\subsection{We investigate the world's largest trees and communities}
\par
\begin{figure*}[tp]
	\centering
	\includegraphics[width=1\textwidth]{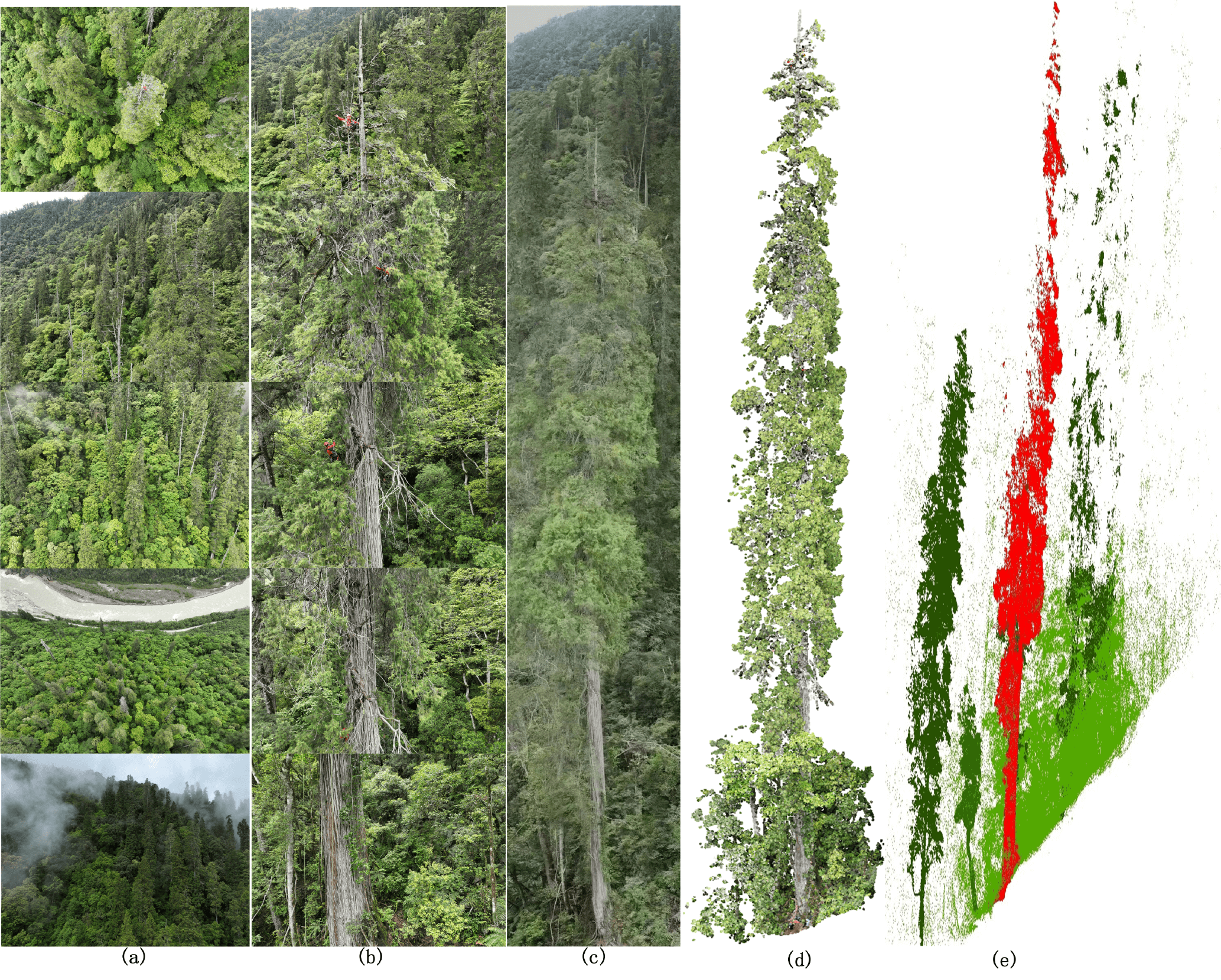}
	\caption{Drone photography and LiDAR scanning of Asia's tallest tree and its giant tree community (a) The tall cypress community in Tongmai Town, Bomi County, represents part of China's most pristine forest; (b) Close-up of different parts of the tallest tree in Asia; (c) Full-length photographs and close-up images of the tallest tree in Asia at 102.3 metres; (d) The tallest UAV-LS point clouds in Asia; (e) TLS point clouds of the tallest tree in Asia.}
	\label{Fig. 2}
\end{figure*}

\begin{figure*}[tp]
	\centering
	\includegraphics[width=1\textwidth]{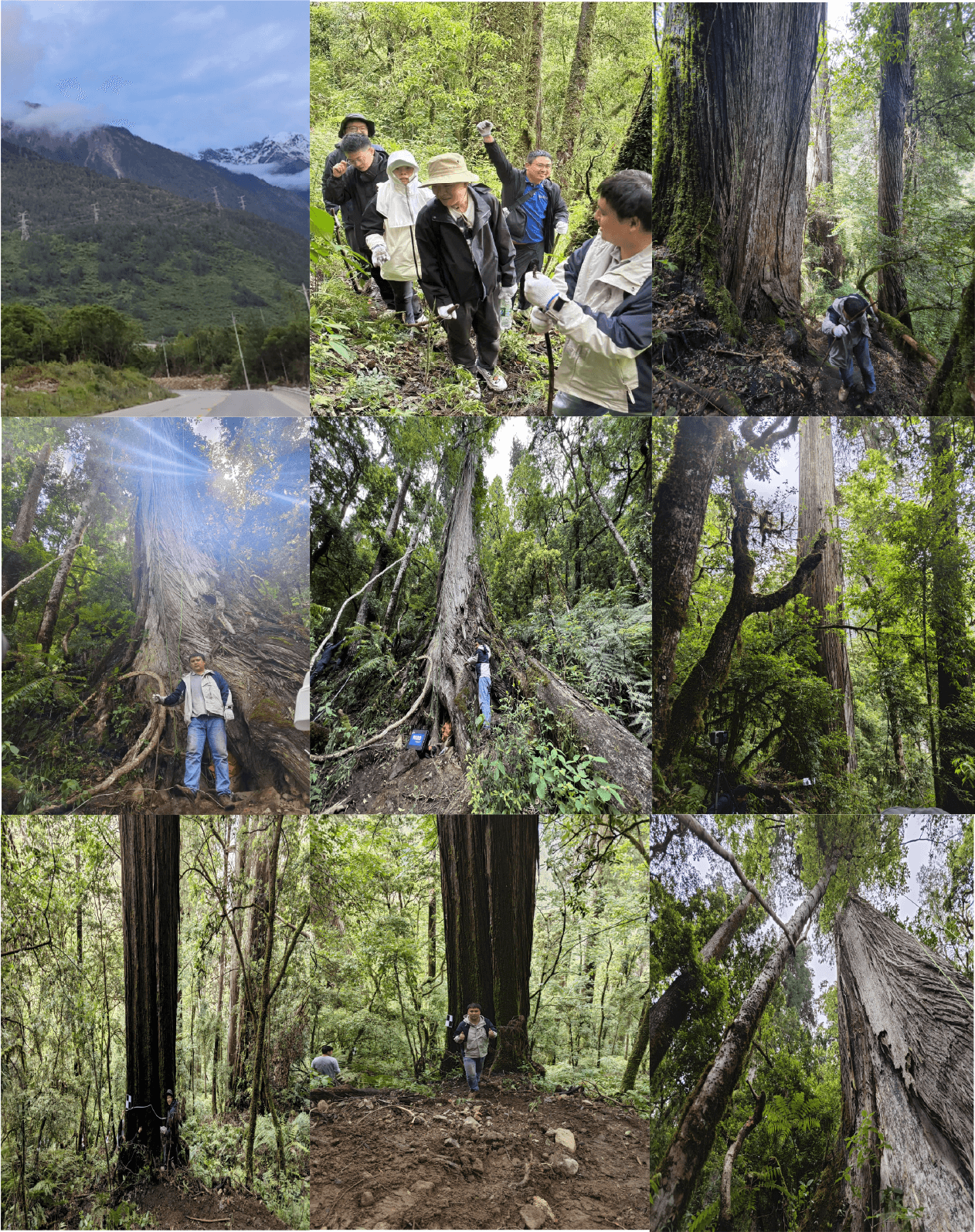}
	\caption{We conducted a field survey of Asia's tallest individual trees and their giant tree communities. The landscape of the fourth world-level giant tree distribution area, from the snow peak of more than 7,000 meters to the Zangpo River Valley of 700 meters, has a vertical range of 7,000 meters, supporting nine vegetation zones and rich and complete biodiversity. One of the authors (Guangpeng Fan) shows the scale of the tallest tree in Asia at the base of the tree and shows the view shot upward within the giant tree community. When we stand in front of these world-level giant trees, we are deeply shocked by their tall and upright posture, feeling like in front of the monument of nature, they seem to be hidden in the primitive forest of ancient giants.}
	\label{Fig. 3}
\end{figure*}

 In 2023, we launched the scientific research project "Weighing the Tallest Tree in Asia", using the unmanned aerial vehicle LiDAR to find the tallest tree in Asia and its forest community (Fig. \ref{Fig. 2}), and conducted a ground sample survey (Fig. \ref{Fig. 3}). The data collected in the field will be used to verify the method proposed in this paper. We found that the top of the trunk of the "tallest tree in Asia" had begun to wither, and estimated that the length of the withered trunk was about 1m at the site, confirming that the tallest tree in Asia had entered a natural mature state, and analyzed that this phenomenon was related to the structure, density, age, site quality and other factors of the giant tree community. At present, from Fig. \ref{Fig. 2}(c),the historical height of the tallest tree in Asia (including the dead trunk) is 102.3 meters, equivalent to the height of a 36-story building \citep*{Time}. The living height (from the base of the trunk to the top of the surviving branches) is 101.2 m. According to the current measurement records, its living height of 101.2 meters is still the second tallest tree in the world and the tallest tree in Asia. The "tallest tree in Asia" belongs to South Tibet cypress, which is only known to exist in parts of the valley of Palong Zangpo and Yigong Zangpo, tributaries of the YTGC, and is a unique plant in China. After comprehensive use of UAV-LS and photogrammetry, ground LiDAR, forest community survey and physical and chemical determination of wood and other technologies, we announced for the first time that the "tallest tree in Asia" weighs 183,139.1 kg, has a carbon stock of 100,365.6 kg, and is more than 1800 years old \citep{ren2024conserving}. The height of other giant trees of the same species in the community is 70 meters and 90 meters, respectively, and the carbon storage reaches 49,270.2 kg and 72797.1kg. Our field survey found that the companion species of Asia's tallest trees, including Huashan pine and Gansu viburnum, reached a height of 30m. The previous work has laid the foundation for this paper, providing strong data support for the discovery of Southeast Tibet-Northwest Yunnan as the fourth world-level giant tree center, and confirming the world-level giant tree individuals and communities represented by the "tallest tree in Asia" as proof of the high productivity of the YTGC region.
\subsection{Data}
\subsubsection{Field data}
Based on the results of the third China National Land Survey and change survey, combined with the "One map" of China's forest resources management and the annual update results of geographical monitoring, we took the forests in southeast Tibet as the scope of ground survey. We reset the fixed plots according to the technical specifications for continuous inventory of national forest resources, set the fixed marks consistent with the third review, and add temporary plots as needed. The sample spacing was 6km×8km, followed the stratified sampling method consistent with the established practice in forest resource survey, and set the sample circle radius of 14.57m and the area of 1 mu (0.0667 hm\textsuperscript{2}). If the sample is located within 14.57 meters of any road, river, or other inaccessible area, it is randomly repositioned along one of the four basic directions. During 2020, we used forest compass, forest survey hand book, RTK, hand-held compass, measuring rope, height rod, DBH circumference ruler, tape ruler, flower rod, reflector and other measuring instruments and tools to carry out plot survey and per wood inspection work. A total of 227 plots were surveyed, and the plant density range was 165-2474 plants /ha. The real-time Kinematic Global Navigation Satellite System (QianxunSI RTK20 equipment) was used to locate the center of the plot with centimeter-level accuracy, using the azimuth and distance to the center of the plot to define the position of individual trees within the plot, and the accuracy was controlled to within 2cm at 60° inclination. In the event of a weakened signal, immediate measures are taken to reposition the GNSS receiver to a more open canopy location, usually 30 meters away. To further improve GNSS positioning accuracy, at least 200 measurements were collected at one-second intervals at each location. Once the exact coordinates of the center of the plot were established, we used a pulsed laser rangefinder and pulsed electronic compass to measure the distance and Angle from the center of the plot to each tree, thus making a geo-reference of all trees with a diameter greater than 5 cm at DBH. In addition to the location of each tree, Records include \emph{Abies fabri}, \emph{Picea asperata Mast}, \emph{Pinaceae}, \emph{Larix gmelinii (Rupr.) Kuzen}, \emph{Pinus yunnanensis Franch}, \emph{Pinus palustris Mill}, \emph{Pinus densata Mast} were measured using a pulsed laser rangefinder. We calculated the average tree height by selecting 10-15 sample trees from the dominant tree species in the main forest layer according to the average DBH. The closure difference of the perimeter measurement is less than 0.5\%, and the length error of the perimeter measurement is less than 1\%. The azimuth Angle and horizontal distance of the sample wood were correctly plotted according to the position diagram of the fixed sample wood, and the error rate of the relative position of the sample wood was less than 3\%. No error is allowed in the number of test rods with a diameter greater than or equal to 8cm, and the error of the number of test rods with a diameter less than 8cm is less than 5\%, and no more than 3 plants. The error is less than 3\% when the tree height is less than 10m, and less than 5\% when the tree height is more than 10m. The measurement error of trees with a DBH equal to or greater than 20cm is less than 1.5\%, and the number of trees with a measurement error of 1.5\% \textasciitilde 3.0\% is not more than 5\% of the total number of trees. For trees with a DBH \textless 20cm, the DBH measurement error is less than 0.3cm, and the number of trees with a measurement error greater than 0.3cm and less than 0.5cm is not allowed to exceed 5\% of the total number of trees. The end result of these rigorous plot survey measures is to ensure that the plot center location produces centimeter-level accuracy to match GEDI and ICESat-2 geographic locations, and to calculate the average tree height of the plot with close spatial resolution.
\subsubsection{UAV-LS data}
We used the Zen L1 LiDAR made by DJI-Innovations and paired it with the Meridian M300 RTK (https://www.dji.com/cn, SZ DJI Technology Co., Ltd.). Zen L1 is composed of LiDAR, mapping camera, three-axis head and other modules, with effective pixels up to 20 million, supports 3-echo scanning mode, and its scanning rate is 480,000 points /s, which can penetrate vegetation and directly detect the surface. We flew drone LiDAR over 6 locations in the study area, covering an area of more than 1,000m\textsuperscript{2}, with each scan lasting more than 20 minutes. We saved the 6 blocks of data into ASPRS standard las files and post-processed them in LiDAR360 (https://www.lidar360.com/, Beijing GreenValley Technology Co., Ltd) software. Firstly, the number of neighborhood points is set to 10 and the multiple of standard deviation is set to 5 by statistical filtering algorithm to separate the noise points. Then, an improved progressive encryption triangulation filter algorithm is used to separate the ground points, where the iteration Angle is 8, the iteration distance is 1.4, and the triangle is stopped when the side length is less than 1. Finally, the non-terrestrial points are finely classified into various categories of point clouds. After removing buildings, towers, wires and other ground objects, the vegetation point clouds are normalized. A digital elevation model and a digital surface model were generated based on the classified data, and the canopy height models of 6 plots were obtained after calculating the difference. To simulate the GEDI waveform data, use the rGEDI (https://github.com/carlos-alberto-silva/rGEDI) treatment after normalization of UAV-LS point clouds. The distance from the center of the plot to each point cloud is calculated, and each point clouds is weighted according to the distance, and then the simulated waveform is generated by Gaussian convolution. Finally, for each analog waveform, different percentiles RH: RH10, RH20, RH30, RH40, RH50, RH60, RH70, RH80, RH90, RH98 are extracted.
\subsubsection{GEDI L1B Waveform data}
The GEDI laser system was installed on the International Space Station (ISS) and flew at an altitude of 415 kilometers, taking measurements along eight ground orbits 600 meters apart. Four ground tracks are generated by two full-power lasers, and another four tracks are generated by a so-called overlay laser with reduced power. The individual lasers are fired 60 meters apart along the orbit, with a lateral distance of about 600 meters between each trajectory. Ten billion waveforms will be measured at the surface with a spatial resolution of 25 m \citep{dubayah2020global}. GEDI offers L1B full-waveform products with a maximum waveform length of 1420 vertical elements and a fixed interval of 1 ns (\textasciitilde 15 cm) between successive echoes.
\par To ensure the accuracy of GEDI canopy height retrieval as much as possible, we first used GEDI Version 2 data. The major improvements in the Version 2 data compared to the Version 1 GEDI data include: (1) improved geolocation of the track segment. (2) Algorithm setting group (SG) selection for each laser shot. We obtained 148 GEDI L1B products (Version 2) distributed by GEDI in the fourth World Giant Tree Center between March 2020 and April 2020. We use the full GEDI L1B waveform as its input and transform it into representative features (identifying peaks and patterns in the waveform) suitable for regression of the wave-shape height measure (98th percentile height) as a proxy for the height of the crown top. We corrected for systematic geolocation errors of GEDI waveforms by correlating sequences (blocks) of GEDI waveforms along a single ground track with simulated GEDI-like tracks of waveforms created from cross-airborne laser scan (ALS) reference data. The process determines horizontal and vertical offsets to maximize the correlation between the on-orbit waveform sequence and the simulated waveform sequence. To improve the GEDI data for further calibration of the model, we use a probabilistic deep learning method based on the set of deep convolutional neural networks (CNN) to process the direct function mapping parameterization from the GEDI L1B waveform to the crown height \citep{lang2022global}. The convolution operation effectively captures the local features and multi-scale structure of waveform data, realizes the translation invariance and hierarchical abstraction of time series data, improves the accuracy and robustness of the model in processing waveform, and completes the effective representation and processing of waveform data.
\subsubsection{GEDI L2A data}
GEDI L2A level 2 data includes two products: GEDI L2A altitude and height measurement and GEDI L2B canopy cover and vertical profile measurement, which are the canopy height and profile indicators at the spot scale. Through waveform data processing, the canopy height and profile indicators are extracted. In this paper, GEDI L2A products are mainly used. Contains elevation and relative height measures derived from the L1B waveform. Each GEDI L2A data contains eight strips, namely BEAM0000, BEAM0001, BEAM0010, BEAM0011, BEAM0101, BEAM0110.BEAM1000, and BEAM1011. Each strip provides basic information about the spot, including the spot's longitude, latitude, ground elevation and relative forest height. The quality\_flag attribute in the GEDI L2A data is used to filter the GEDI L1B data. The data whose value is 0 is filtered out, the waveforms that may be invalid are deleted, and only the valid waveforms whose value is 1 are retained. In this paper, 183 GEDI L2A data collected between March and June 2020 were used to produce a training dataset matching GEDI L2A with 2,033,951 footprint light spots. The data represents a waveform return measure for each 25-meter diameter GEDI footprint, and the footprint data is geolocation.

\subsubsection{ICESat-2 data}
NASA launched ICESat-2, a new space-based laser altimeter satellite carrying the Advanced Terrain Laser Altimeter System (ATLAS), in September 2018. ATLAS uses a laser emission method of 3 to 6 beams (each pair of laser beams consists of a strong laser beam and a weak laser beam, and the energy ratio of strong and weak beams is 4:1). The ICESat-2 data product announced by NSIDC is divided into four levels. A total of 22 products, respectively ATL00 \textasciitilde ATL21, all use HDF5 file format to store data. The data products related to ground elevation and canopy height mainly include ATL03 and ATL08. ATL03 is a global positioning photonic data product that includes photon time, latitude, longitude and elevation information.
\par We used ATL03 data from Level-2 in our study, The national snow and ice data center (https://nsidc.org/data/icesat-2/Data-sets) to download the during March 2020 to August 2020 in article 88 of the research region ATL03 data products for subsequent analysis, three strong beams of each ATL03 are reserved for extraction of forest canopy height. ATL03's photon point clouds retains a large amount of noise distributed over land, near the surface, or below the surface. These noises may come from the reflection of clouds, the reflection and scattering of sunlight or other sources, which will have a great impact on the extraction of canopy information. The DBSCAN algorithm does not need to specify the number of clusters in advance, and is suitable for unknown terrain features. It can effectively process noise and outliers in data, and classify them into noise points and outliers, which can improve data quality. We used DBSCAN for noise removal and photon recognition on ATL03 data, setting the radius to 8.9 and the minimum number of points to 11. Photon classification method is used to set a 10-meter step length for the photon along the track. The highest point of the step length is taken as the top photon of the canopy, and the lowest point is taken as the ground photon. By subtracting the height of the upper point from the height of the lower point to obtain the canopy height value, we finally obtained 782,268 effective canopy top photons in the study area.
\subsubsection{Sentinel-2 data}
Sentinel-2 is a group of high-resolution multispectral imaging satellites that currently consists of two identical satellites: the first is Sentinel 2A, launched in 2015, and the other is Sentinel 2B, launched in 2017. The revisit cycle of one satellite is 10 days, and the complementary use of two satellites can achieve a five-day revisit cycle. Each Sentinel-2 satellite carries a multispectral instrument (MSI) that can cover 13 spectral bands with ground resolutions of 10 m, 20 m and 60 m. Four of these bands provide a ground sampling distance (GSD) of 10 meters in the blue, green, red, and near infrared (NIR) regions. With its near-infrared and short-wave infrared bands, Sentinel-2 is specifically designed to capture vegetation features, among other things.
We in the European space agency (https://dataspace.copernicus.eu/) to download the hidden in January of 2020, covering 84 pieces of cloud cover is less than 20\% of the southeast and northwest Yunnan, pass the SNAP Sen2Cor radiation calibration and atmospheric correction in lower atmosphere reflectivity L2A level data. In addition, using images with less cloud cover, five L2A class images with less than 45\% cloud cover from February to August of the same year were extracted to supplement the coverage of the study area. Each tile has a step length of 110km in the UTM WGS84 projection, with a 10km overlap with adjacent tiles. From each file, we extracted spectral bands of 10 m resolution (B02, B03, B04, B08), 20 m resolution (B05, B06, B07, B8A, B11, B12), and 60 m resolution (B01, B09), for a total of 1008 bands. Load all bands of a tile into the Bend set of QGIS according to tile classification, create a Mosaic using Mosaic of band sets, set Data outside the study area to 0, and use Clip raster bands to cut in batches according to vector boundary data in the study area. Repeat the above steps to crop all tiles, resulting in a total of 3142 (jp2 format) files. All images were cropped according to the boundaries of the vectorized study area, and the 12 bands of each image were synthesized. After the raster file is obtained, the Mosaic operation is carried out to obtain a complete multi-spectral image of the fourth world-level giant tree center through color homogenization. We loaded the image of the study area into QGIS (3.28.11), and cut the Sentinel-2 data of the whole southeast Tibet into 84 files according to the vector boundary of each tile extracted.
\section{Methods}
\subsection{Data fusion of spaceborne LiDAR and Sentinel-2}
The combination of surface borne LiDAR and optical image is more effective than a single data source to improve the prediction accuracy of canopy height. First, we select the latitude, longitude and altitude indicators extracted from GEDI and ICESat-2, using a total of 2,832,791 footprints. To solve the problem of spatial heterogeneity of footprint data generated by two space-borne LiDAR, we use the characteristic distribution of ICESat-2 and GED to maintain the same expectation and variance. Any band is selected from the subset Sentinel 2 image for height rasterization, that is, the height index of GEDI and ICESat-2 through the footprint is associated with the image pixel respectively. We chose the blue band B02 with a spatial resolution of 10 meters and a wavelength range of about 459-509 nanometers (nm), which is mainly used to measure visible light reflection from the surface. It is sensitive to the reflection of blue light, so it can provide information about the blue characteristics of the surface. The latitude and longitude corresponding to the relative altitude indexes of GEDI and ICESat-2 were matched with the Sentinel images. The width, height, spatial resolution, position and other information are extracted from the B02 band to be rasterized, and the latitude and longitude corresponding to the height indexes of GEDI and ICESat-2 are converted into a grid consistent with the image shape using an affine matrix. The specific range and position of the satellite-borne LiDAR footprint in the image are determined by the row number of the pixels, and the data fusion between the satellite-borne LiDAR and Sentinel-2 is completed after repeated operations.
\subsection{ Pyramid receptive field depth-separable Convolutional Neural Network (PRFXception)}
\subsubsection{The architectural design concept of PRFXception}
Considering that the fourth world level giant tree distribution area consisting of southeast Tibet and northwest Yunnan has unique geographical characteristics in the world. There are global and local differences in spatial distribution of primary forests, and simply introducing conventional CNN or deep separable convolutional neural network may lead to more uncertainties. To map the height of the primary forest canopy in the fourth world-level giant tree distribution area in high resolution, we propose a deep learning method called PRFXception. Pyramid Receptive Field Depthwise Separable Convolution (PRFXception) is a customized pyramid receptive field DSC. PRFXception efficiently coupled multi-dimensional sensitive field features and was designed specifically for deep learning modeling driven by multi-source Earth observation data from spaceborne LiDAR (GEDI, ICAESAT-2) fusion multi-spectral images (Sentinel-2). Inspired by the research work of Xception and Nico Lang, we make PRFXception fully learn the data features after the fusion of spaceborne LiDAR and spectral images, perform pixel level semantic segmentation of remote sensing images, and restore the tile size to the original size \citep{chollet2017xception}. To classify a pixel, an image block around the pixel is used as input to the PRFXception.
\par Here, we use PRFXception to regression canopy height from data fused from GEDI, ICESat-2, and multispectral Sentinel-2 images. Compared with traditional Convolution operations, PRFXception inherits and retains the features of DSC, reducing the number of parameters while reducing the computational complexity. PRFXception can further stack layers without causing training difficulties or gradient disappearance. A deep convolution operation is first performed on each input channel, each of which has a corresponding convolution kernel. After the deep convolution, a convolution kernel of 1x1 (point-by-point convolution) is used to combine the channel information. Compared with traditional convolution, the number of parameters is less, the computational complexity of the model is reduced, the space and channel features are captured better, and the training and inference process is accelerated \citep{fayad2021cnn,fu2024automatic}. DSC improves the ability of the model to express and generalize complex patterns, and reduces the risk of overfitting, especially when the data set in this paper is small compared to the global scale data. Although DSC helps to improve the ability of the model to process features at different scales, the purpose of introducing multiple receptive fields in this paper is somewhat different. The importance of multiple receptive fields in neural networks is reflected in that it can effectively capture and fuse features of different scales, which is particularly critical when processing multi-source Earth observation data.
\par We introduce the multi-field mechanism to improve the capability of PRFXception to capture features of different scales: (1) Improve the flexibility of feature capture: different scale field can capture features of different levels. Small receptive fields are better at capturing details (such as edges, textures, etc.), while large receptive fields are better at capturing broader features (such as overall shape, context, etc.). Multiple receptive fields allow PRFXception to focus on both local detail and global structure, thereby improving the understanding of multi-source Earth observation data fusion in this paper. (2) Enhanced feature expression richness: By combining features of different scales, neural networks can generate richer and more comprehensive feature expression. This multi-level feature combination provides a more comprehensive information basis for high regression and subsequent classification or recognition. (3) It improves the generalization ability of the model: the multi-receptive field structure makes the model more adaptable to different data changes, including scale, shape and texture, etc. This adaptability enables the model to maintain better performance in the face of diverse data. (4) Optimizing computational efficiency: Compared with simply increasing the network depth or width, multiple receptive fields provide a more efficient way to enhance the feature extraction capability of the network. By using parallel multi-scale convolution kernel, the performance of the model can be improved without significantly increasing the computational burden. (5) Better processing of complex scenes: In complex scene understanding tasks, such as regression tasks, image segmentation, object detection, or natural language processing, multiple fields of perception can help the model better understand the relationship between different objects and their position in the whole scene.
\par On the basis of PRFXception inheriting the advantages of DSC, we add multi-scale parallel convolution to the network, which contains convolution cores of different sizes. PRFXception's pyramid receptive field can capture features at different scales simultaneously - small convolution checks are more sensitive to detail, while large convolution cores capture a wider range of context information. The addition of multi-scale parallel convolution may increase the computational complexity and parameters of the model. We carried out careful design and balance to avoid excessive increase in computational burden, and used fusion techniques such as concatenation and summation to achieve effective fusion of parallel convolution layers to integrate features of different scales. PRFXception's parallel convolutional layer design is coordinated with deep separable convolution to maintain model efficiency. 
\par The novelty of PRFXception includes: (1) Scale diversity, with multiple receptive fields enabling the network to observe input data through different sized receptive fields, capturing multiple levels of information from microscopic details to macroscopic structure. (2) Feature fusion: Different scale receptive fields can extract different levels of features. In rasterized image processing, a small receptive field may capture details such as edges or textures, while a large receptive field can understand the overall shape of the object or the context of the scene. (3) Strong adaptability: By combining multiple receptive fields, the network can better adapt to different types and scales of data and enhance its generalization ability. The differences between PRFXception and the traditional single receptive field neural network model architecture include: (1) Receptive field size: In the traditional single receptive field network, all convolutional layers usually have a fixed size receptive field, such as 3×3 or 5×5 convolutional kernel. This limits the ability of the network to capture features at different scales. (2) Information processing: Networks with a single receptive field may not work well in processing complex data containing multi-scale information because they cannot capture both detail and overall information at the same time. (3) Network structure: The introduction of multiple receptive fields usually requires a more complex network structure, such as parallel convolutional layers or the introduction of hollow convolutions, while the network structure of a single receptive field is relatively simple.
\subsubsection{Architecture of PRFXception}
PRFXception realizes pixel-level analysis of Sentinel-2 satellite images by DSC and residual connection. The PRFXception architecture is mainly divided into three parts, using a series of convolution, separation convolution layers, and point convolution blocks. PRFXception considers multiple sized receptive fields at the same layer, using multiple branches with different sized kernels to capture multi-scale information. Each layer of PRFXception has four parts, including 1×1 convolution, 3×3 convolution, 5×5 convolution, 7×7 convolution, and 3×3 maximum pooling. In the convolution operation, the convolution layer with different kernel size and the pooling layer are used in the way of padding='SAME' to ensure that the output feature image has the same size, so that the final result is combined on the Depth channel. Multiple convolution kernels are used to extract the information of different scales of the image and finally fuse it to obtain better representation of the image. Convolution kernel of different sizes (1×1, 3×3, 5×5, 7×7) are used respectively on the same layer. Convolution kernal of different sizes provide receptive fields of different scales, and features of different scales are extracted from the same layer. Features of different scales can be combined to produce more features than using a single size. The PRFXception is weighted to sum the responses of different scale receptive fields so that the neurons can synthesize input information from multiple scales (Eq. (1)).
\begin{equation}
	F(x)=\sum_{i=1}^kf_i(x)
\end{equation}
\par In the convolutional neural network of a certain layer of PRFXception, there are a set of convolutional or pooled kernels of size k, which correspond to different scale receptive fields. Let $f_i(x)$ represent the response of the convolutional kernel or pooled check of the i scale to input x, and $F(x)$ represent the response of the layer's neurons to input x, then the response of the layer's neurons can be expressed as a summary of the responses of all scales.
\par First, the entry part of the model is built by PointwiseBlock, which accepts the number of input channels in\_channels, and uses a set of convolutional layers with channel numbers [128, 256, num\_sepconv\_filters] as the feature extractor. This part is mainly responsible for the initial feature extraction and dimension transformation of the input data. Second, by building a series of separated convolution blocks sepconv\_blocks. These blocks further process and extract features to better capture complex patterns and relationships in the input data, and separating convolutional structures helps improve the model's perception of local features. Finally, output is generated from three 1×1 convolutional layers predictions, variances and second\_moments. These convolution layers are used to generate the model's predictions, variances, and second moments, respectively. PRFXception preserves the residual connection of Xception to mitigate the vanishing gradient problem(Chollet, 2017). These connections are between DSC blocks, facilitating smooth propagation of gradients and making the network easier to train. All blocks are residuals, and their inputs are also passed by skipping joins over that block, and are added to the output activation map so that the whole block learns additional residuals for the identity map. Skipping connections facilitates learning of very deep networks by creating shortcuts that prevent error gradients from disappearing before reaching earlier layers. PRFXception reduces the number of channels through 1×1 convolution to gather information, and then carries out feature extraction and pooling of different scales, and then superposes the features after obtaining the information of multiple scales. PRFXception makes heavy use of jump joins, does not use activation functions and BatchNorm on the jump join side, and it aligns the number of channels by 1×1 convolution. The combination of convolution blocks and separation convolution blocks improves the ability of the model to characterize multi-source remote sensing data by stacking and combining these blocks. Through the final convolutional layer, multiple outputs are generated, and the network architecture that meets the requirements of the forest canopy height regression task in this paper is developed.
\par
\begin{figure*}[tp]
	\centering
	\includegraphics[width=1\textwidth]{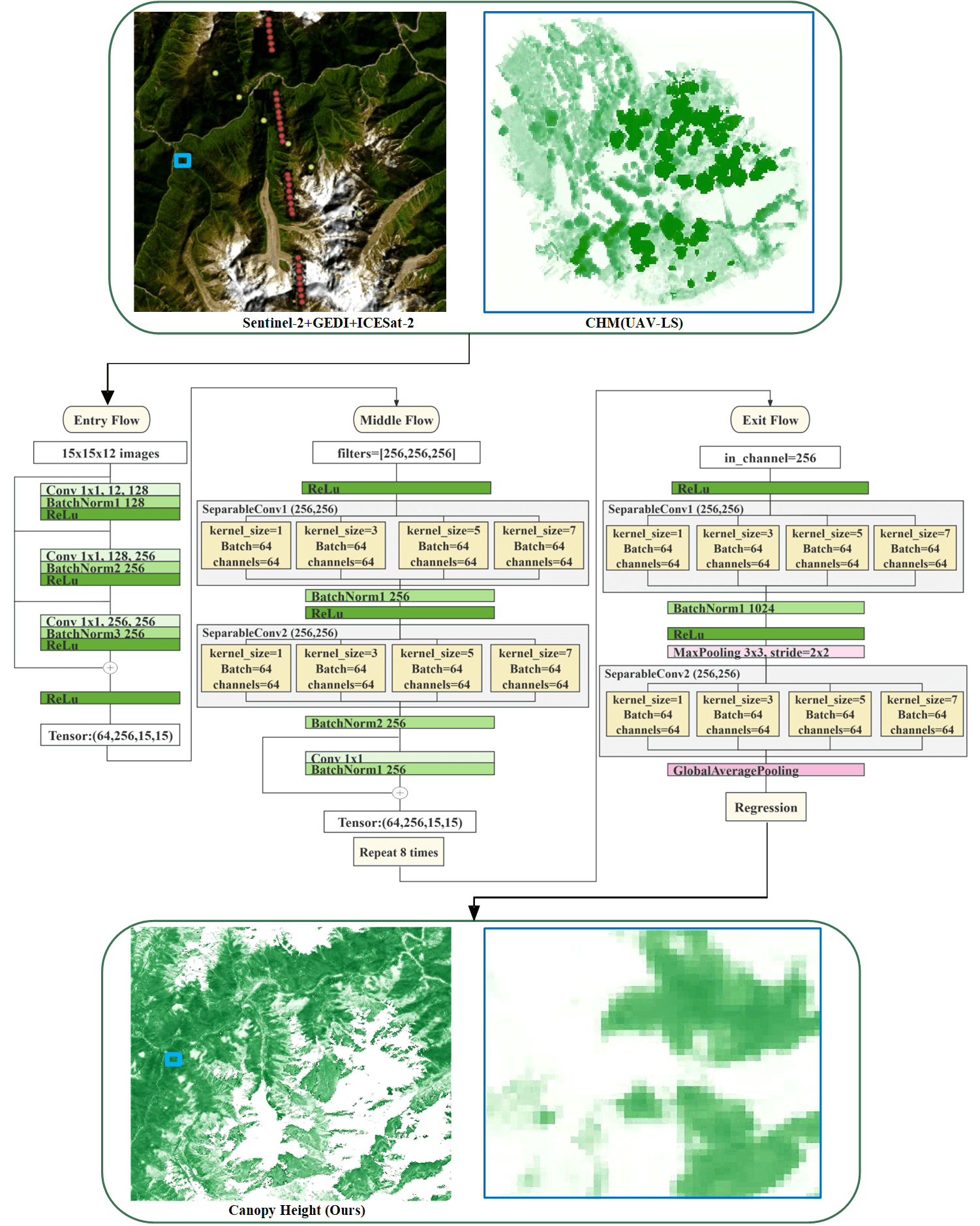}
	\caption{Architecture diagram for PRFXception.}
	\label{Fig. 4}
\end{figure*}

 Previous studies have shown that the spatial information around pixels is very important to the performance of image classification algorithms. In many cases, a single pixel may not provide enough information to make an accurate prediction, especially if there is noise or spectral mixing. The multi-scale parallel calculation of PRFXception takes into account the spatial-spectral characteristics of each pixel of the fused spaceborne LiDAR image, the intensity of the light reflected, absorbed and emitted by the original forest with the unique geographical environment of the world at different wavelengths in 13 bands, and the physical and chemical properties of the forest are integrated in the convolution process(Fig. \ref{Fig. 4}). In the local extraction of spectral features, the convolution layer is used as a filter to slide along the wavelength axis of the spectrum, and the pyramid receptive field architecture effectively filters and screens the spectral features at a deeper level by stacking multiple convolution layers.
	\FloatBarrier
\subsection{Model training}
Canopy height retrieval is a pixel-by-pixel regression task. We trained the regression model end-to-end in a supervised manner, meaning that the model learned to convert raw image data into spectral, color, and texture features to predict crown height. We split the data set into non-overlapping training sets and test sets. In the training set, a 10\% subset is randomly selected as a validation set to monitor the optimization process. Thus, CNN parameters fit 90\% of the training set. Validation sets are not used for optimization, but are used to monitor the loss of unseen data (cumulative prediction error). We train convolutional neural networks by sparse supervision, i.e., selectively minimizing losses only at pixel locations where GEDI and ICESat-2 reference values are present (Formula 2). Before entering the 15-channel data cube into the CNN, each channel is normalized to a standard normal distribution using the channel statistics from the training set. The learnable parameters (weights) are randomly initialized and used as a loss function by mean square error (MSE) in the neural network regression task to improve the prediction accuracy of the canopy height by PRFXception. By normalizing the fit by adding an optional l2 penalty (" weight decay ") to the parameter, the total loss function is minimized as follows Eq. (2):
\begin{equation}
	Loss=\dfrac{1}{N}\sum_{i=1}^{N}(f(x_i)-y_i)^2+\lambda\dfrac{1}{W}\sum_{j=1}^{W}(w_j)^2
\end{equation}
Model $f$ and its weight $w_i$ (including constant deviation for each kernel), input intensity $x_i$, reference canopy height value $y_i$, and prediction $f(x_i)$ at pixel $i$. $N$ and $W$ represent the number and weight of samples, respectively. The training process consists of 200 cycles with 5000 iterations per cycle. To optimize the training process and update the parameters, an adaptive version of Adam optimizer based on stochastic gradient descent (SGD) was selected for 1,000,000 iterations of training, the limit of maximum gradient value was 1000, and the regularization weight was l2\_lambda.
Combining momentum optimization and adaptive learning rate adjustment strategies. First, the first-order moment (mean) and second-order moment (uncentered variance) of momentum are defined Eq. (3)(4):
\begin{equation}
	m_t=\beta_1\cdot m_{t-1}+(1-\beta_1)\cdot g_t
\end{equation}
\begin{equation}
	v_t=\beta_2\cdot v_{t-1}+(1-\beta_2)\cdot g_t^2
\end{equation}
where, $g_t$ is the gradient of the current iteration, $m_t$ is the first moment of the gradient, similar to the moving average of the gradient, $v_t$ is the second moment of the gradient, similar to the moving average of the gradient squared, and $\beta_1$ and $\beta_2$ are the exponential decay rates, usually set to a value close to 1. Then, correct the deviation of the first and second moments (Eq. (5)(6)):
\begin{equation}
	\widehat m_t=\frac{m_t}{1-\beta_1^t}
\end{equation}
\begin{equation}
	\hat{v}_t=\frac{v_t}{1-\beta_2^t}
\end{equation}

where $t$ represents the number of current iterations. Finally, the parameters are updated using the corrected first and second moments as defined Eq. (7):
\begin{equation}
	\theta_{t+1}=\theta_t-\frac{\eta}{\sqrt{\hat{v}_t}+\epsilon}\cdot\widehat{m}_t
\end{equation}
$\theta_t$ is the argument on the $t$ iteration, $\eta$ is the learning rate, and $\epsilon$ is a small number that prevents the denominator from being zero. By comprehensively considering the first and second moments of the gradient, as well as momentum and adaptive learning rate adjustment, the parameters can be updated more effectively, the convergence rate of the model can be accelerated, and the model performs well on the dataset and tasks in this paper.
In each iteration of the optimizer, a batch size of 5 (batch\_size) is used for each training iteration to maintain the training stability of the model while maintaining computational efficiency. The partial derivative (gradient) of the model parameters is determined by backpropagation, using the chain rule to calculate the numerical gradient. Set the learning rate to 0.0001, the milestone list of the learning rate to 400-700, patch\_size to 15, and slice\_step to 1. It is reduced in subsequent iterations to speed up the convergence process of model parameters. We use the MultiStepLR learning rate scheduler to adjust the learning rate of the neural network during training, and reduce the size of the learning rate after different training stages and fixed time steps to fine-tune the parameter update step of the model. By gradually reducing the learning rate, the model becomes more stable in the late training period and finds a better solution near the global optimal point, which is beneficial to avoid drastic fluctuations and overfitting in the training process, and improves the generalization performance. The network parameter with the lowest validation loss is used in the final model and then applied to the retained test set to evaluate model performance. The selection and setting of this method aim to provide an efficient and accurate training framework for canopy height regression based on DSC neural networks.

\subsection{Random cross validation}
To verify the performance of the model on different data subsets, so as to more accurately evaluate the generalization ability of the model. In our task, we perform random 10x cross-validation to evaluate the performance of the model on different data subsets to improve the reliability of the evaluation results. The original data set was randomly divided into 10 mutually exclusive subsets, and then 10 experiments were conducted to train the model and evaluate the performance on the test set. One subset at a time is selected as the test set, and the remaining nine subsets are used to train the 10 CNN models from scratch. The selection and partitioning are repeated until each subset acts as a test set, i.e., the complete dataset has been retrained 10 times. The overall performance indicator is calculated by averaging the outcome indicator of 10 experiments so that each fused data appears only once in the test set as an indicator for the final model evaluation.
\subsection{Geographic cross validation}
\begin{figure*}[htbp]
	\centering
	\includegraphics[width=1\textwidth]{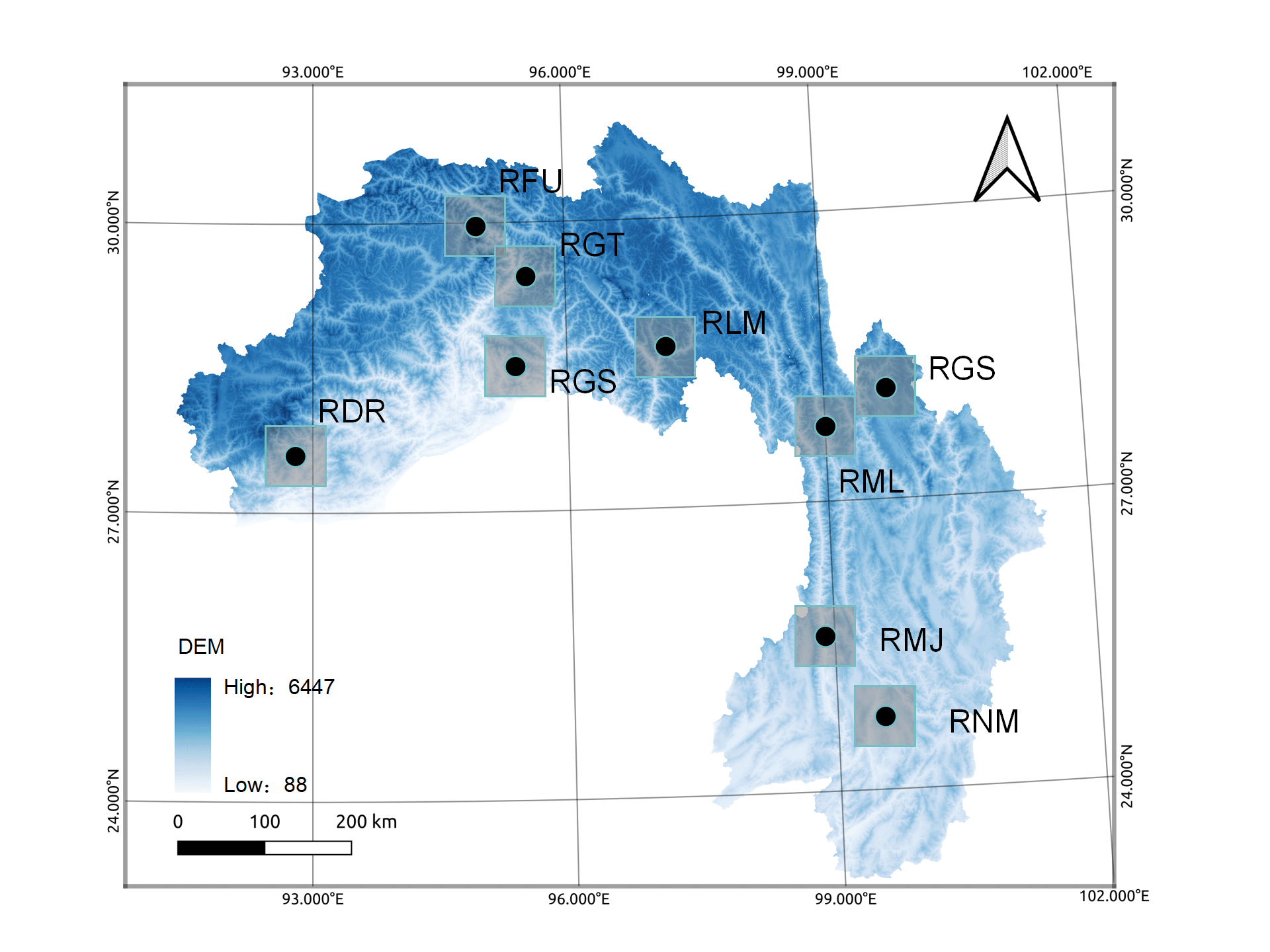}
	\caption{Geographical cross-validation region selection distribution map}
	\label{Fig. 5}
\end{figure*}

We use geographic cross-validation to evaluate the ability of the model to generalize to different regions. Cross-geographic regions (with fairly similar climates and biomes) were tested by cross-validation between their respective regions. We verify the geographical transferability of the model by dividing geographical regions of the world-level giant tree centers mainly in southeast Tibet. We trained the model from scratch in the southeast Tibet region and kept all available samples from the northwest Yunnan geographical region for testing. In this way, samples from the northwest Yunnan region will not be seen during the training process. With geographic cross-validation, we can assess whether the model overfits the training location or whether it supports common features such as making predictions at unseen locations. We conducted geographical cross-validation for 5 regions in southeast Tibet and 4 regions in northwest Yunnan respectively (Fig. \ref{Fig. 5}), which will produce 5 times and 4 times cross-validation. We conducted two types of geographic cross-validation. In the first, we trained two separate regions, southeast Tibet and northwest Yunnan, and then cross-predicted their respective regions. The second method is to cross-verify the five selected regions in southeast Tibet and four regions in northwest Yunnan. All but one area is trained, and the remaining test area is predicted so that the training data near the test area is not visible during the training period.
\subsection{Evaluation index}
We use the root mean square error (RMSE) as the primary error measure to quantify the deviation between the predicted height and the reference height. The mean error (ME) is used to measure the performance of a training model evaluated over an invisible test area. For completeness and to allow comparisons with other studies, we also report the mean absolute error (MAE).
\begin{equation}
	RMSE=\sqrt{\dfrac{1}{N}\sum_{i=1}^N(f(x_i)-y_i)^2}
\end{equation}
\begin{equation}
	MAE=\frac{1}{N}\sum_{i=1}^{N}\mid f(x_i)-y_i\mid 
\end{equation}
\begin{equation}
	ME=\dfrac{1}{N}\sum_{i=1}^N(\widehat{y}_i-y_i)
\end{equation}
where $f(x_i)$ represents the predicted value of the model set, and $y_i$ represents the reference value at sample $i$. According to this definition, a positive ME (mean square error) means that the value predicted by the model is higher than the true value of the reference data.

\section{Results and discussion}
\label{sec5}

\subsection{Canopy height regression}
\label{subsec5}


\subsubsection{Validation of reference data for GEDI and ICESat-2}
\begin{figure*}[tp]
	\centering
	\includegraphics[width=1\textwidth]{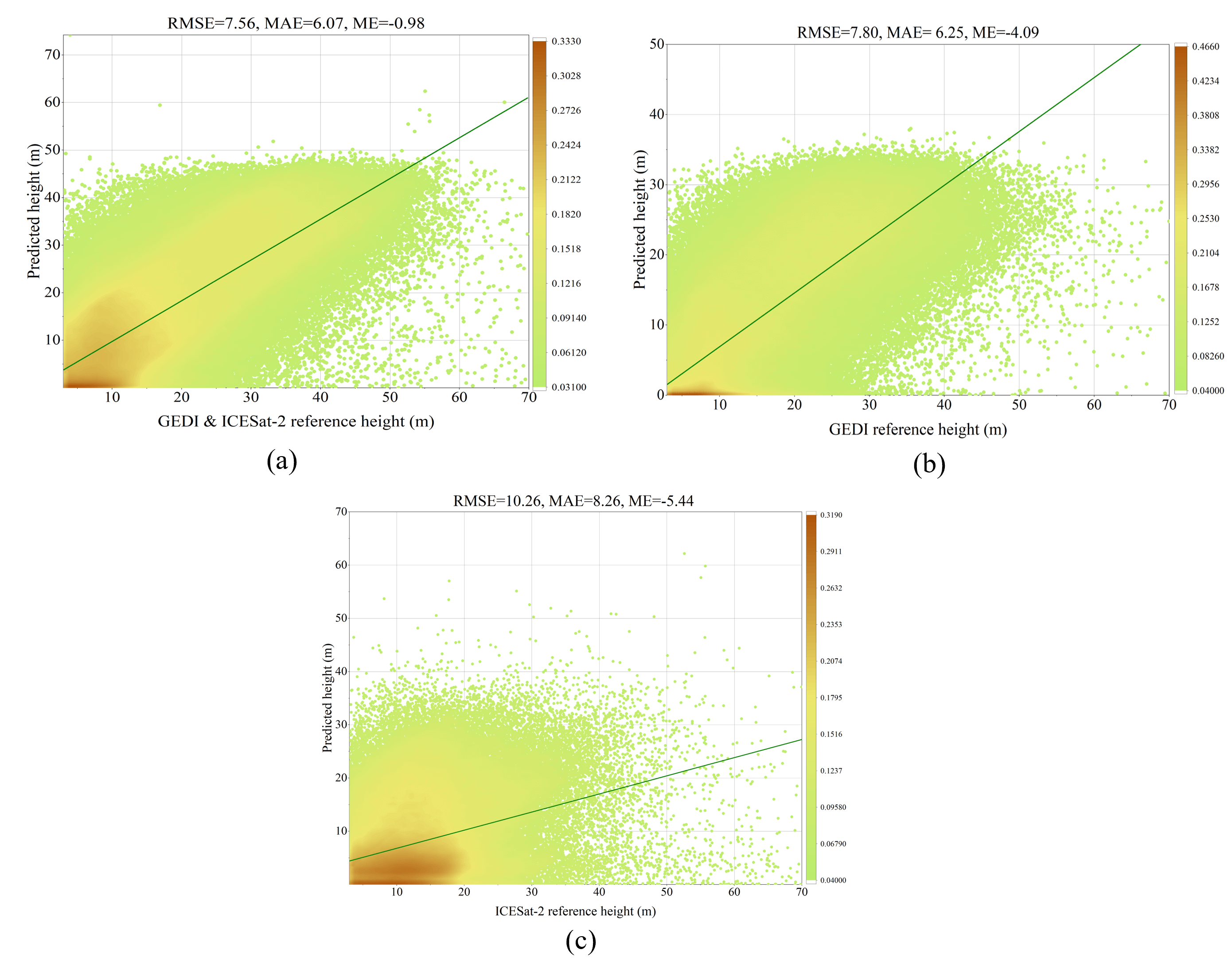}
	\caption{Point density plots of reference and predicted canopy height values obtained. (a) ICESAT-2 \& GEDI; (b) GEDI; (c) ICESat-2}
	\label{Fig. 6}
\end{figure*}

\begin{figure*}[tp]
	\centering
	\includegraphics[width=1\textwidth]{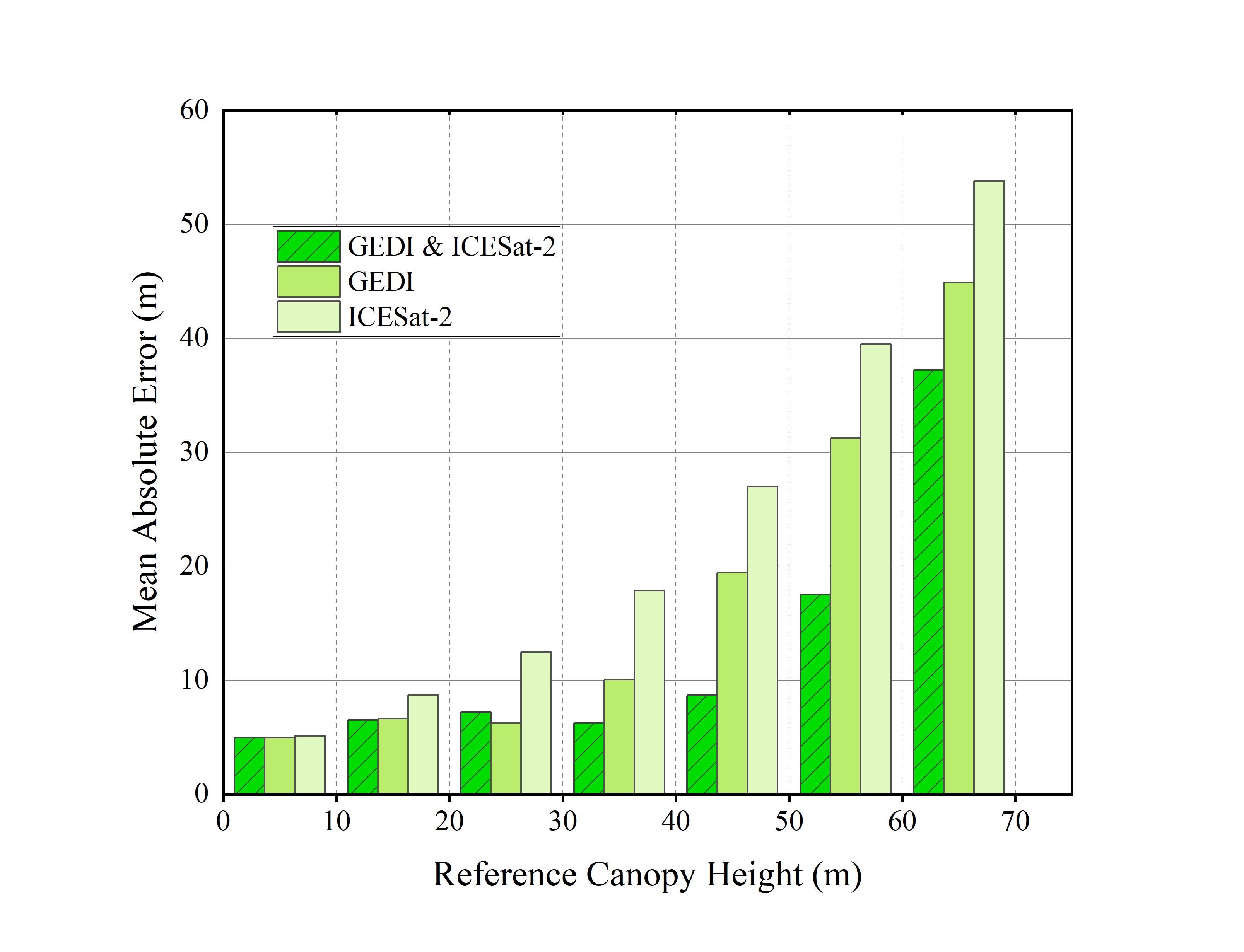}
	\caption{Mean absolute error per 10 m canopy height intervals among 3 groups of data}
	\label{Fig. 7}
\end{figure*}

\begin{figure*}[tp]
	\centering
	\includegraphics[width=1\textwidth]{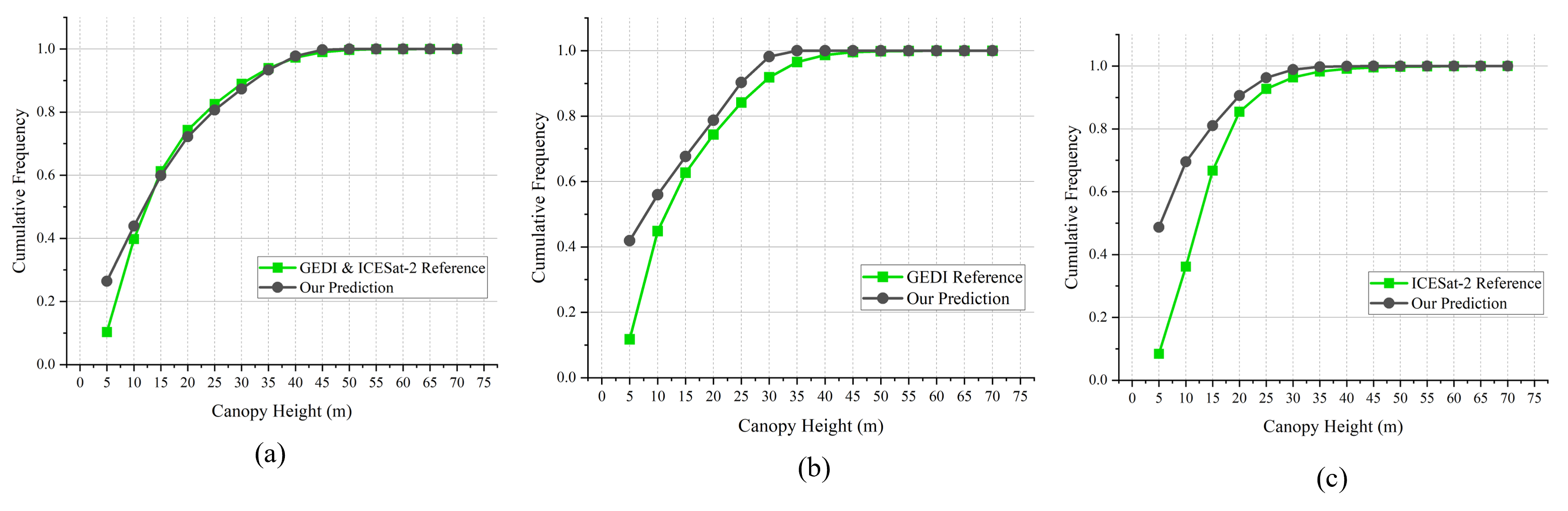}
	\caption{Cumulative distribution of canopy height prediction. (a) ICESAT-2 \& GEDI; (b) GEDI; (c) ICESat-2}
	\label{Fig. 8}
\end{figure*}

\begin{figure*}[tp]
	\centering
	\includegraphics[width=1\textwidth]{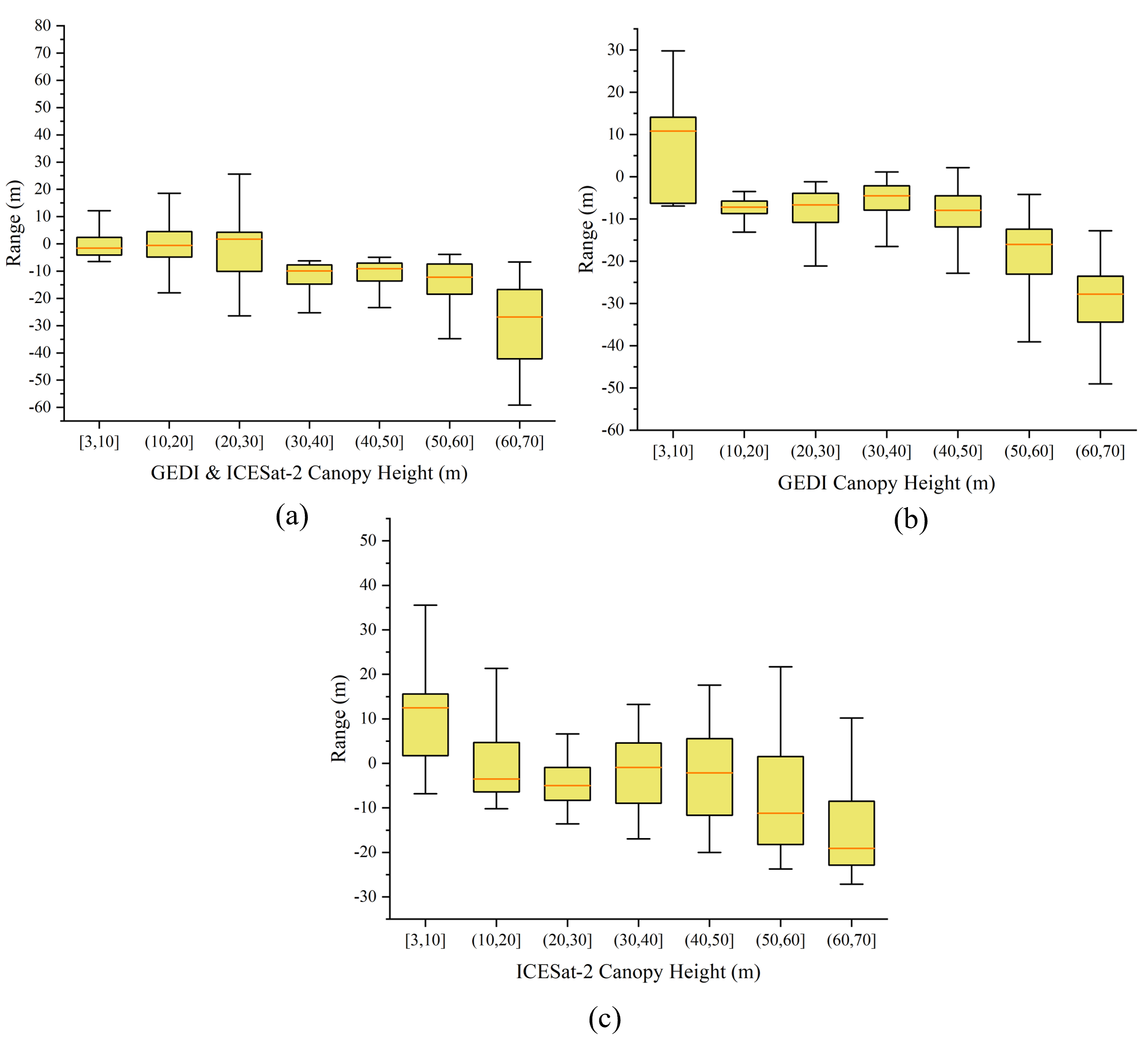}
	\caption{Forest canopy height corresponds to residual box diagram. Negative residual indicates that the predicted height is lower than the reference value. (a) ICESAT-2 \& GEDI; (b) GEDI; (c) ICESat-2}
	\label{Fig. 9}
\end{figure*}

We use three different combinations of training data as sparse supervised samples to train PRFXception and predict height separately. The data combination types were ICESat-2, GEDI, and ICESat-2 fusion GEDI, respectively, and sample labels were used as reference data to verify the accuracy of high regression. The results show that the canopy height predicted by PRFXception is in good agreement with the reference data. By comparison, it can be seen that the combination effect of ICESat-2 \& GEDI is better, $RMSE$=7.56m, $MAE$=6.07m, $ME$=-0.98m, indicating that the model has a high level of fitting ability. The fitting line is close to the 1:1 line, indicating that the predicted forest canopy height is not saturated. Most of the data points are concentrated near the fitting line, but with the increase of the canopy height, the dispersion of the predicted canopy height also increases gradually. This means that for taller trees, the model's prediction accuracy decreased slightly. We observed that the forest canopy height is underestimated after the canopy height exceeds 50m. In the verification results using ICESat-2 data and GEDI data alone, RMSE was 7.80m and 10.26m, MAE was 6.25m and 8.26m, and ME was -4.09m and -5.44m, respectively. It can be seen that ICESat-2 fusion GEDI data has better accuracy in mapping pixel-level continuous forest canopy height than single spaceborne LiDAR data(Fig. \ref{Fig. 6}).
A more detailed view of the predicted performance in different height categories is given in Fig. \ref{Fig. 7}. By dividing the canopy height into height intervals of 10m, we found that the accuracy of canopy height regression decreased with the increase of height, that is, MAE gradually increased. This confirms the trend that the lower the height and density of vegetation, the more accurate the prediction. We observed that the model performed better in the height range below 50 meters, and the canopy height predicted by GEDI combined with ICESat-2 had lower MAE in almost every interval than the data source alone.
The cumulative frequency distribution of the reference altitude of GEDI \& ICESat-2 combination, GEDI reference altitude, ICESat-2 reference altitude and corresponding predicted altitude are respectively shown in the three graphs in Fig. \ref{Fig. 8}. It can be seen from the comparison of the three maps that GEDI combined with ICESat-2 as the reference value and the model prediction value, the cumulative frequency increases with the increase of canopy height, and our prediction is closely consistent with the reference data combined with GEDI and ICESat-2. Using the two types of data alone, our predicted height value is higher than the reference value. In Fig. \ref{Fig. 8}, the two curves are very close, which indicates that our model prediction has a high agreement with the reference data, especially in the high canopy height interval, the model prediction and the reference data almost coincide. According to our prediction (Fig. \ref{Fig. 8}), about 70\% of the trees' canopy height is less than 20m, and about 15\% of the land in southeast Tibet-northwest Yunnan is covered by trees of 30m or more.
From Fig. \ref{Fig. 9}, we can see that for the lower canopy height (10-30m), the residual distribution is relatively tight and the median is close to 0, which indicates that the prediction is relatively accurate in this range. Boxes with canopy height in the range of 50-70m show a wider residual distribution, which indicates that the variability of the prediction in this height interval increases and the prediction accuracy may decrease. At higher canopy heights (\textgreater 60m), the box is very short, there are fewer data points, and the predicted value may be lower than the reference value. The median of the GEDI \& ICESat-2 combined data set is closer to zero, indicating higher accuracy across the entire canopy height range.
\subsubsection{Ground survey reference data verification}
\begin{figure*}[tp]
	\centering
	\includegraphics[width=1\textwidth]{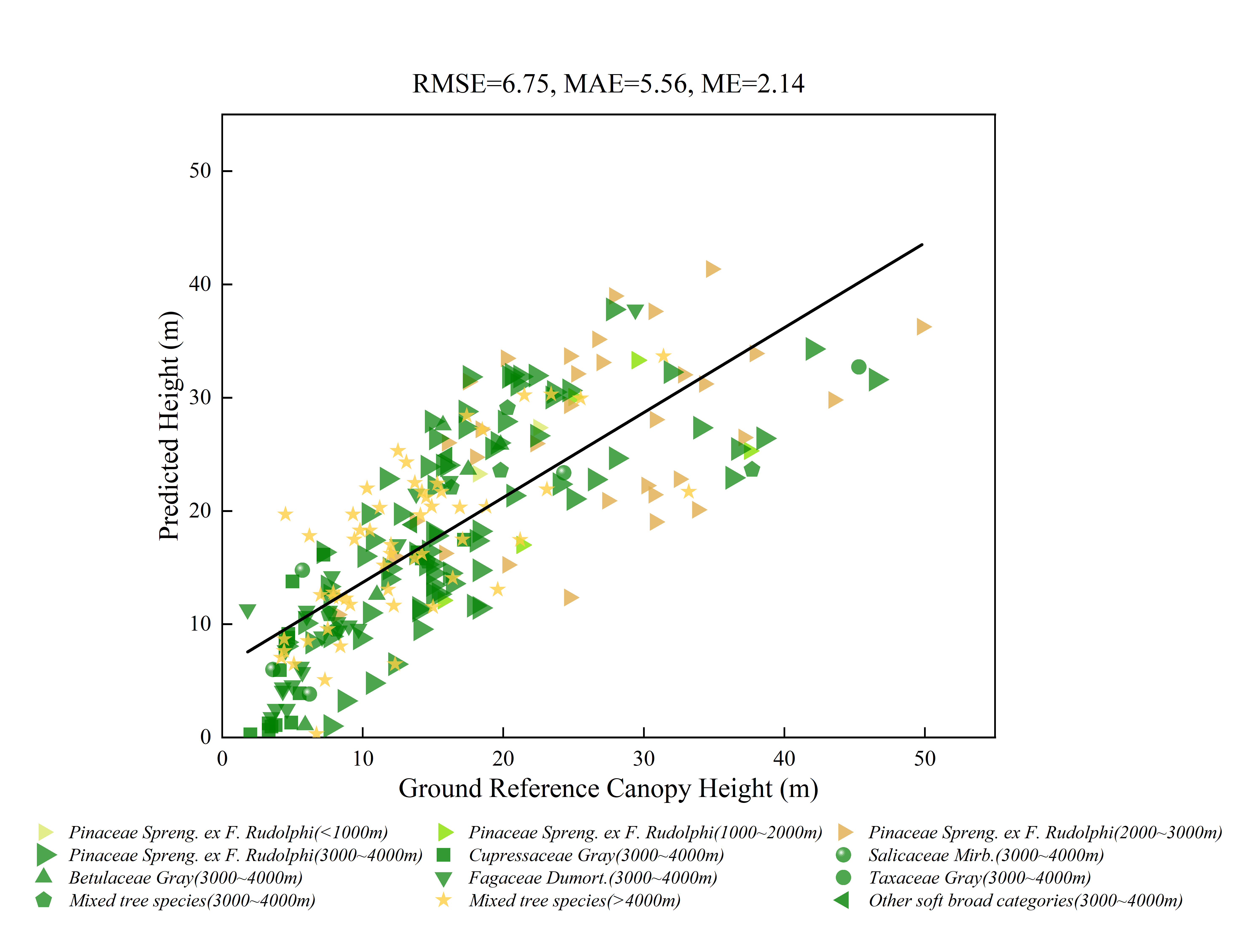}
	\caption{The predicted canopy height is compared with the measured canopy reference height in the permanent plot}
	\label{Fig. 10}
\end{figure*}

We used 227 permanent plot survey data as reference values to verify the accuracy of PRFXception's prediction of canopy height. The predicted height for each sample is obtained from the output of the PRFXception. To visually show the consistency of the reference data with the predicted results, we produced a scatter plot to show the relationship between the ground survey and the predicted height. RMSE=6.75m, MAE=5.56m, ME=2.14m. In Fig. \ref{Fig. 10}, the colors represent different elevations and the shapes represent different tree types. We can find higher trees distributed in the range of 2000-4000m above sea level; In areas with elevations over 4000m, the height of the canopy is mostly within 20m.
Considering the location matching accuracy of the sample data set and the satellite observation data set which are not completely synchronized, the reliability of the accuracy assessment results using field measurements may be limited. In an assessment using ground survey data, we observed a slight underestimation of forest canopy height in forests over 40 m in height. Comparing the model results with the field measurements, the predicted forest canopy height slightly overestimated the reference canopy height in the forest with the canopy height less than 20 m.
\subsection{Model accuracy evaluation}
\begin{figure*}[tp]
	\centering
	\includegraphics[width=1\textwidth]{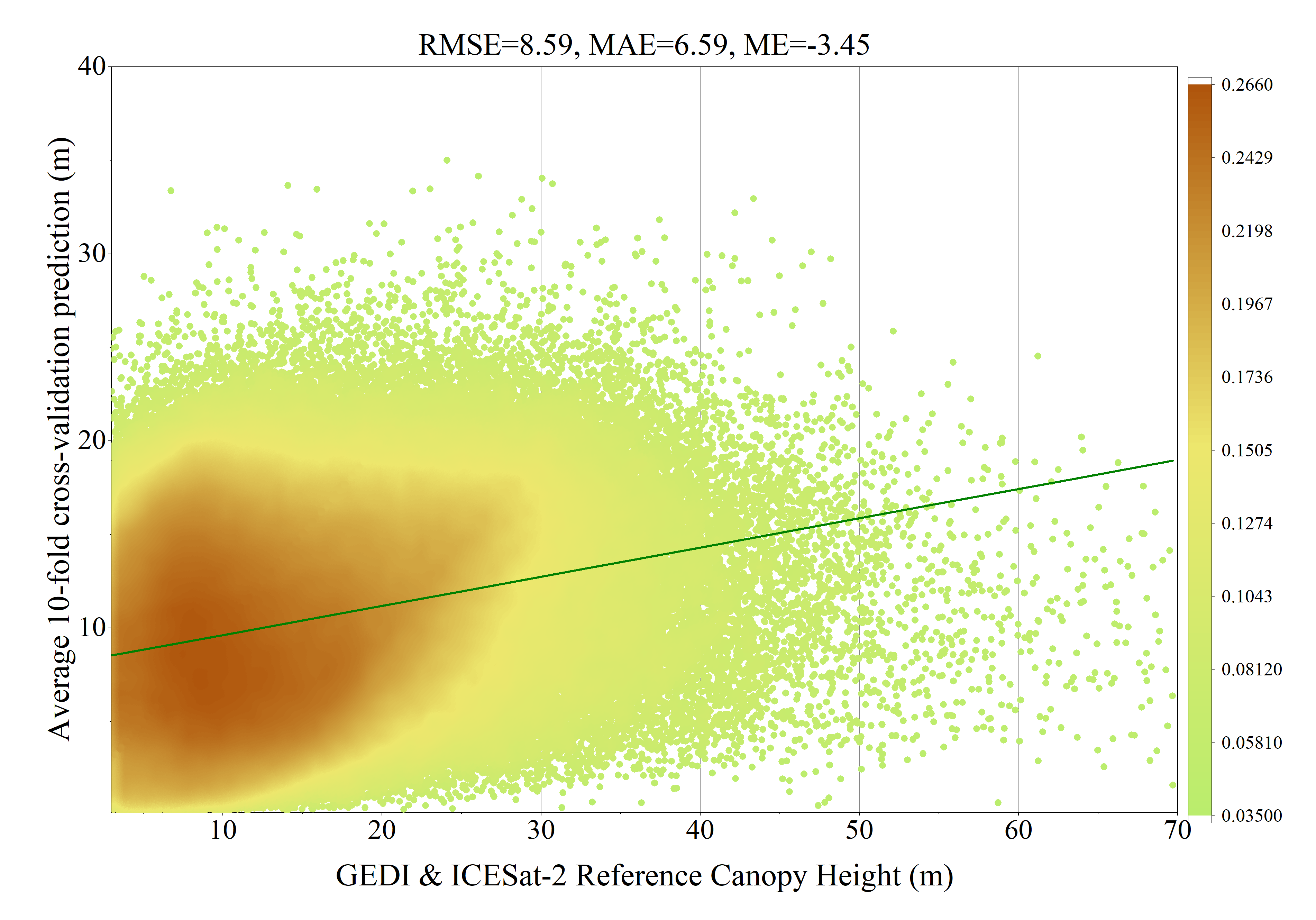}
	\caption{10-fold cross-check validation evaluates the performance of PRFXception}
	\label{Fig. 11}
\end{figure*}

We perform random 10x cross validation to measure whether model performance remains constant for different training/test splits. Evaluate and select the performance of the PRFXception model through cross-validation. We divide the dataset into 10 mutually exclusive subsets, 9 of which serve as the training set and the remaining 1 as the validation set, and repeat 10 times, with each subset used as the test set once. To obtain a more robust estimate of model performance. The experiment produced a total of 10 models, and we counted the prediction results of each model separately. Finally, the model performance was evaluated using the root-mean-square error RMSE, MAE, and ME (Fayad et al., 2021). In this study, random cross-validation on this dataset produced RMSE=8.59m, MAE=6.59m, ME=-3.45m (Fig. \ref{Fig. 11}), indicating a slight underestimation error compared to the GEDI \& ICESat-2 reference value, showing a good agreement between the predicted value and the reference height.
\subsection{The effect of spectral bands}
To fully understand the effects of Sentinel-2’s different spectral bands on mapping primary forest canopy heights, we trained and tested 5 band combinations. They include GSD of 10m in visible band RGB (B02, B03, B04), only 10m in near-infrared band N (B08), visible light and near-infrared 10m band combination RGBN (B02, B03, B04, B08), short-wave infrared, visible and near-infrared band combination woRGBN (B01, B05, B06, B07, B08a, B09, B10, B11, B12), all 13 band combinations.
\begin{figure*}[tp]
	\centering
	\includegraphics[width=1\textwidth]{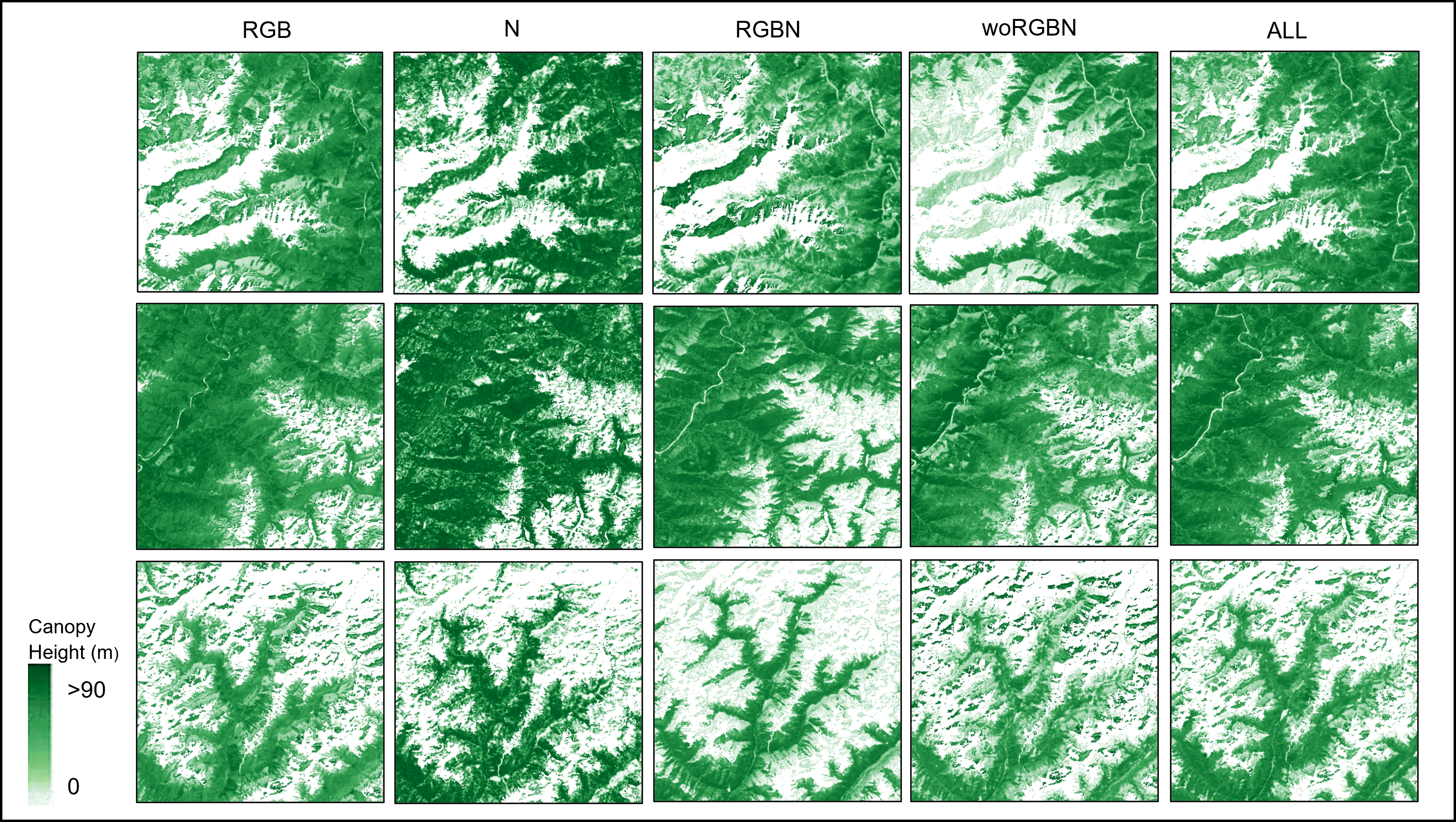}
	\caption{Comparison of 5 band combinations}
	\label{Fig. 12}
\end{figure*}
\par Fig. \ref{Fig. 12} shows the results of using different spectral bands to draw the forest canopy height at the same location. It can be found that the canopy height drawn by the data set containing RGB band provides more abundant information contouring, texture, tone and shape, while only N or the introduction of short-wave infrared will reduce the canopy height rendering effect.
\begin{table}[]
	\caption{RMSE, MAE, ME values generated by different band combinations}
	\begin{tabular}{cccc}
		\hline
		band      & RMSE  & MAE   & ME    \\ \hline
		RGB       & 6.32  & 5.32  & 0.43  \\
		N         & 11.91 & 10.07 & 5.02  \\
		RGBN      & 6.85  & 5.77  & 0.22  \\
		woRGBN    & 6.34  & 5.36  & -1.51 \\
		All bands & 6.54  & 5.51  & -0.73 \\ \hline
	\end{tabular}
	\label{tb1}
\end{table}

\par We changed the number of input band channels, training them separately while keeping all other configurations of the PRFXception unchanged. Table \ref{tb1} shows the performance evaluation of canopy height regression accuracy for different data sets (RGB, N, RGBN, woRGBN, all-band) using Sentinel-3 satellite images. We can find that the RGB band alone performs better, with the lowest RMSE and MAE values, but the mean deviation of the single-band NIR (N dataset) is higher. For results in the same region, the performance of ALL and RGBN is similar, indicating that the high-resolution band carries most of the relevant information, while the other nine bands up-sampled to 10 GSD contribute little. The all-band method is not significantly better than the RGB method in accuracy. There is a tendency to underestimate, which may be due to the introduction of extra noise in too many bands, or some bands are not very important for estimation of canopy height. Near the top of the height range, the 20m and 60m bands did not significantly improve regression and seemed to have a negative effect. This is consistent with the observation that for tall trees, textural features are more beneficial, but further investigation of more geographic diversity is needed to confirm this fact. RGBN also achieves lower ME under very low vegetation. Further studies are needed to analyze the reason, which may be related to the implicit spatial smoothing of spectral information when sampling up to 10m in low-resolution bands. Even slightly improved results can be obtained when using only RGB or woRGBN instead of all bands. However, assessing the significance of this result based on one test region is not statistically significant and further testing of different regions is required. Perhaps the clearest message of this study is that the use of near-infrared rays alone is not enough, and when we predict the height of the primary forest canopy, we cannot obtain enough multi-spectral satellite images, and selecting data with RGB band will be a potential alternative.
\subsection{The influence of multiple receptive field characteristics}
One reason for using CNN as a regressor is that we hypothesize that with Sentinel-2's resolution up to 10 meters, individual trees can be as large as a pixel footprint, and texture features may play an important role. By stacking many convolutional layers with filter cores of different scales, such as 1×1, 3×3, 5×5, 7×7, etc., PRFXception learns to pick up patterns of texture features in a larger sensory field centered on each pixel, if they are highly correlated with the canopy. To check if this is necessary, we set the space size of all convolution cores to 1×1, limiting the network to focus only on the spectral distribution of each individual pixel, preventing it from seeing any texture or spatial context. Otherwise, the CNN architecture (number of layers, depth of channel at each layer) is the same, so the network only derives its prediction at that pixel from the spectral intensity of a given pixel, but still allows the same degree of nonlinearity in the map. These lower-level activations are further combined into a larger receptive domain, resulting in more task-specific features. Should this situation combine spatial textures with longer time series and more complex multi-temporal representations than our simple median to further improve the prediction? The fact that we can get very reasonable predictions even from a single image also suggests that textures may be able to compensate for time redundancy to some extent, and vice versa.
\par
\begin{figure*}[tp]
	\centering
	\includegraphics[width=1\textwidth]{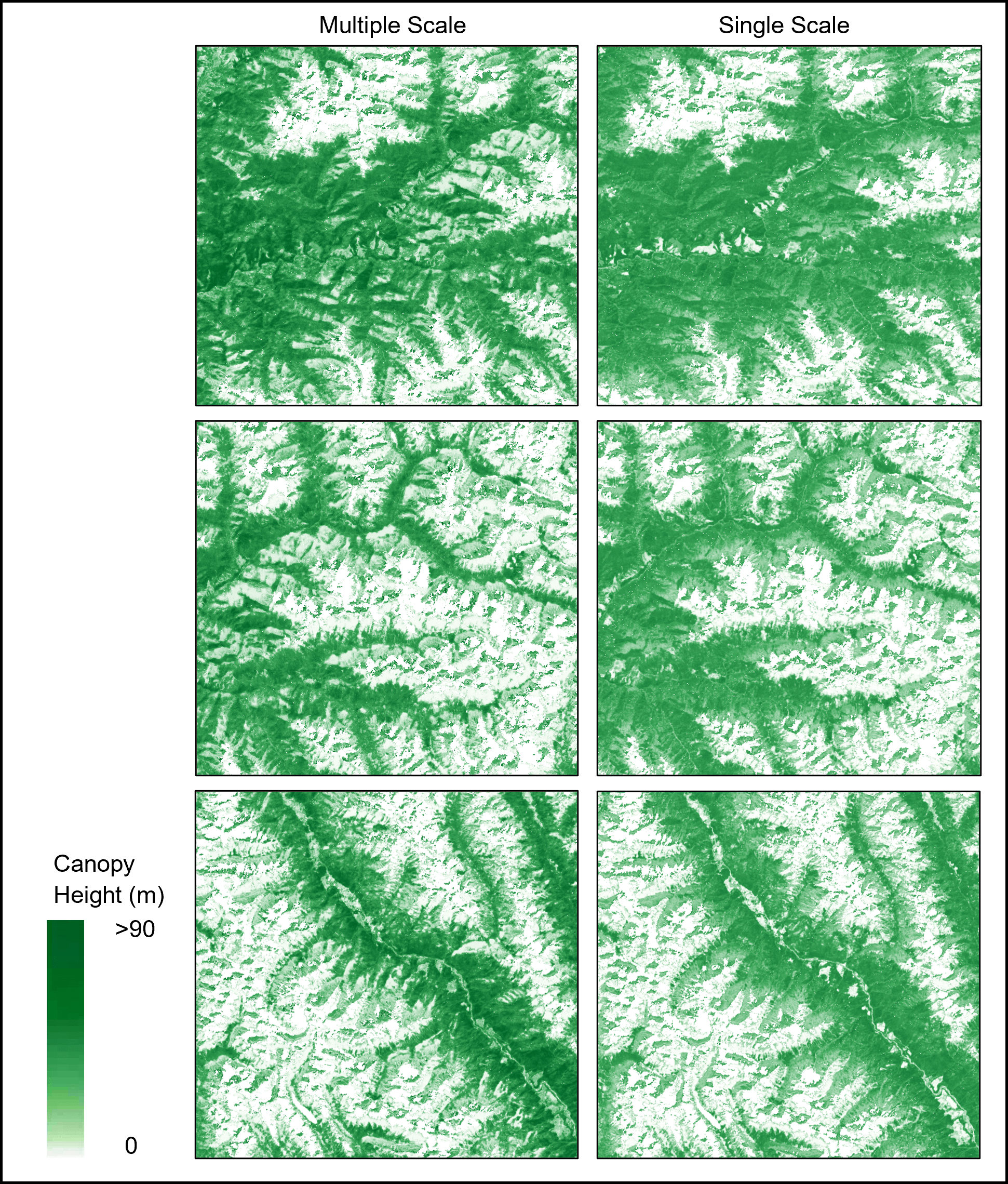}
	\caption{Pyramid receptive field (1×1, 3×3, 5×5, 7×7) compared with single receptive field (1×1)}
	\label{Fig. 13}
\end{figure*}

 Our PRFXception was specifically customized and developed for remote sensing applications in search of world-level giant trees, and the biggest difference from regular CNNS is a deep regression network dedicated to predicting the height of primary forest canopy. PRFXception combines features of different scales to obtain more features than a single size. The multi-scale parallel structure not only enhances the ability of the network to capture features of different scales, but also improves the flexibility and robustness of the model. Fig. \ref{Fig. 13} represents the height contrast of the canopy in the same location drawn by the multi-receptive field model and the single receptive field model. The height detail of the multi-receptive field drawn in the first column is clearer, especially in the forest edge and spatial structure, which indicates that the introduction of the multi-receptive field mechanism can make the model better capture the complexity of terrain and space. In contrast, the height of the single receptive field is relatively blurred, and the forest edges, textures, and detailed features are not as obvious. At the same time, we can find that multiple receptive fields also seem to have advantages in capturing high canopy areas, which can be concluded from the color distribution and change of pixel values in the map.
\par From Fig. \ref{Fig. 13}, we can see that PRFXception has a more refined spatial rendering mode, because the pyramid receptive field can capture more subtle features, the painted canopy height transitions and changes in different locations are smoother and smoother, and the texture information of the terrain is richer. In addition, details are shown more precisely, such as small forest areas or the edges of the tree canopy. In contrast, the height of the single receptive field is relatively rough, the changes are relatively abrupt, and the context information generated by the larger structure may be missed. Compared with the traditional CNN architecture, our approach greatly reduces the computational burden of the model. By parallel convolution kernels of different sizes, the model can capture features of multiple scales at the same layer, which is very beneficial for understanding the spatial complexity of forest canopy. For example, a smaller convolution kernel can capture detailed information, while a larger convolution kernel can capture broader context information. In a “wider” network, the generalization ability of the model generally increases due to the reduction in model depth, which helps reduce the risk of overfitting. The advantage of PRFXception is that it can better handle the regression task of multi-scale target objects, and can extract and fuse multi-scale feature information on different levels of feature maps to obtain more spatial spectrum and texture information. Our model has the ability to perceive and generalize objects at different scales, and is more effective in retrieving objects with different scale variations.

\begin{table}[]
	\caption{Comparison of the accuracy of PRFXception in five regions of southeast Tibet with the predicted value of a single receptive field}
	\begin{tabular}{cccccccc}
		\hline
		&     & \multicolumn{3}{c}{Multiple scale} & \multicolumn{3}{c}{Single scale} \\
		&     & RMSE      & MAE       & ME         & RMSE      & MAE      & ME        \\ \hline
		1 & RDR & 8.26      & 7.03      & 0.47       & 10.17     & 8.42     & 7.34      \\
		2 & RFU & 9.83      & 8.36      & -2.94      & 12.55     & 10.48    & -8.87     \\
		3 & RGS & 8.15      & 6.86      & -1.74      & 13.80     & 11.88    & -10.34    \\
		4 & RGT & 7.40      & 6.78      & -2.9       & 14.24     & 15.32    & -14.33    \\
		5 & RLM & 7.02      & 5.96      & -3.06      & 10.17     & 8.59     & -7.39     \\ \hline
	\end{tabular}
	\label{tb2}
\end{table}
\par We used the model trained by multiple receptive field and single receptive field to predict five different sites in southeast Tibet, and the accuracy of the multi-receptive field model was better than that of the single receptive field model in different regions of southeast Tibet (Table \ref{tb2}). The RMSE of the multi-receptive field in the five regions ranged from 7.02m to 8.26m, the MAE from 5.96m to 8.36m, and the ME from -3.06m to 0.47m. The single receptive field in the five regions had RMSE ranging from 10.17m to 14.24m, MAE ranging from 8.42m to 15.32m, and ME ranging from -14.39m to 7.34m. On the whole, the prediction error of multi-receptive field model is lower than that of single-receptive field model, which emphasizes the potential of PRFXception model in improving the prediction accuracy of geographical features of different regions in southeast Tibet. This means that our model can provide information about the reliability of the predicted results. This is very valuable for predicting forest canopy height, especially in areas with topographic complexity and vegetation diversity, such as southeast Tibet. Our method provides reliable canopy height prediction in this region, which provides an important basis for national nature reserve policy makers to plan for primary forest management and protection. By understanding the distribution of canopy height, forest managers can better understand the structure and function of forest ecosystems and formulate corresponding conservation and management strategies.
\par Previously Nico Lang used a convolutional layer with a 3×3 filter kernel to pick up textures with a larger receptive field centered on each pixel \citep{lang2019country}. To overcome the high computational cost and overfitting problems caused by large convolutional layers in the construction of deep convolutional neural networks, we designed a unique PRFXception model. Parallel multi-scale convolutional operations are introduced in a custom deeply separable convolutional neural network architecture. This design allows the network to process convolution cores of different sizes in parallel at the same level, thus widening the width of the network rather than the depth. The receptive fields of multiple sizes (1×1, 3×3, 5×5, 7×7) are considered in the same layer, dynamically adjusting rather than fixing a particular size. Multiple branches with different sizes of kernel are used to capture multi-scale information, and multiple convolution kernels are used to extract image information of different scales, and finally better image representation is obtained after fusion.
\subsection{Cross-validation of geographic generalization}
Deep neural networks have a large model capacity, that is, enough trainable parameters to adapt to the details of the training data distribution. To verify whether PRFXception is overfitted in a particular training area or suitable for areas other than the training area, we conducted a geographical training/test segmentation experiment on the study area to study geographical generalization. We used the model trained in southeast Tibet to verify the forest canopy height in northwest Yunnan, so as to evaluate the generalization ability of the model. Geographic cross-validation plays an important role here, not only to reveal the applicability of the model to new regions, but also to help identify and correct the limitations of the model in different geographical conditions. In the experiment, all samples from the test area are used for evaluation, and the model is trained from scratch without seeing any samples from that test area. The geographical generalization results of the fourth world level megalopolis are given in Table \ref{tb3}.
\par We carried out geographical cross-verification of 5 sub-regions in southeast Tibet and 4 sub-regions in northwest Yunnan. The selected areas in southeast Tibet are Shannan Nature Reserve (RDR), Lulang National Forest Park (RFU), YTGC National Nature Reserve (RGS), Medog County National Nature Reserve (RGT), and Zayü Nature Reserve (RLM). The selected areas in northwest Yunnan are Gaogongshan National Nature Reserve (RML), Nujiang Grand Canyon (RMJ), Qinghai Hua National Wetland Park (RNH) and Shangri-La Grand Canyon (RNM). We first trained the models of the two overall regions of southeast Tibet and northwest Yunnan, and then predicted the selected sub-regions respectively (Column 1 of Table 3). After that, we carried out cross-verification in southeast Tibet and northwest Yunnan. Train all but one subregion for prediction so that training data near the test area is not visible during training. The experiment will yield 5x and 4x cross-validation, respectively (Column 2 of Table 3).
\par As can be seen from Table 3, on the whole, RMSE ranges from 6.15m to 9.77m, MAE ranges from 5.17m to 8.35m, and ME ranges from -0.55m to 3.75m in southeast Tibet. In northwestern Yunnan, RMSE ranges from 5.43m to 7.19m, MAE ranges from 4.60m to 6.20m, and ME ranges from -3.96m to -0.20m. It can be seen that the overall prediction effect in southeast Tibet is lower than that in northwest Yunnan, which may be due to the greater elevation difference in southeast Tibet. For southeast Tibet, PRFXception showed higher accuracy in RLM area, with lower RMSE, MAE and ME, and better prediction effect. The overall error deviation of PRFXception in northwest Yunnan is low, and the uncertainty in RNM is relatively high, which may be due to the relatively high canopy or elevation of Shangri-La Grand Canyon compared to other regions. There is a low deviation in the RNH area, which is likely to belong to the national wetland Park with a low canopy height, so the accuracy of the model prediction is higher.
\begin{table*}[]
	\caption{The accuracy values of overall model prediction and residual cross-validation for five regions in southeast Tibet and four regions in northwest Yunnan}
	\begin{tabular}{cccccccc}
		\hline
		&     & \multicolumn{3}{c}{Column 1}                 & \multicolumn{3}{c}{Column 2}                  \\
		&     & \multicolumn{3}{c}{Overall model prediction} & \multicolumn{3}{c}{Residual cross-validation} \\
		&     & RMSE          & MAE          & ME            & RMSE           & MAE           & ME           \\ \hline
		\multirow{5}{*}{Southeast Tibet}  & RDR & 9.77          & 8.35         & 3.75          & 8.14           & 6.94          & 0.35         \\
		& RFU & 8.19          & 6.96         & 0.94          & 14.28          & 11.74         & -3.35        \\
		& RGS & 9.08          & 7.77         & 3.39          & 8.07           & 6.79          & -1.75        \\
		& RGT & 7.40          & 6.32         & 0.93          & 8.33           & 7.02          & -3.13        \\
		& RLM & 6.15          & 5.17         & -0.55         & 8.34           & 7.00          & -3.98        \\ \cline{2-8} 
		\multirow{4}{*}{Northwest Yunnan} & RML & 5.86          & 4.99         & -0.20         & 5.90           & 5.02          & 0.28         \\
		& RMJ & 5.73          & 4.85         & -0.28         & 6.08           & 5.10          & -2.21        \\
		& RNH & 5.43          & 4.60         & -0.47         & 6.76           & 5.73          & -0.62        \\
		& RNM & 7.19          & 6.20         & -3.96         & 6.13           & 5.24          & -0.88        \\ \hline
	\end{tabular}
	\label{tb3}
\end{table*}
Geographic cross-validation is able to test models beyond the limitations of their training environment and make predictions across regions. We tested the prediction results of the model trained on the dataset of southeast Tibet in northwest Yunnan, RMSE=10.09m, MAE=8.79m, ME=6.59m. In the whole northwest Yunnan region, the RML, RMJ, RNH and RNM predictions were exceeded by the root-mean-square error, and the mean error (ME) was greater than 10-fold. This may be because the canopy height in northwest Yunnan is mostly low, and the prediction deviation caused by the training model in southeast Tibet is slightly larger. When using Northwest Yunnan as the test set, PRFXception already sees some data from Northwest Yunnan, while section "4.1" takes all available data from Southeast Tibet-Northwest Yunnan as the test set. However, we did not observe any significant performance differences. Overall, the experiments in this section show that the proposed model generalizes fairly well to geographic regions not seen in training and does not overfit to specific (training) regions. Through geographic cross-validation, we can find the performance difference of the model in different regions, so as to provide a basis for revealing the influence of factors such as climate, soil and vegetation type on the prediction of forest canopy height in different regions. Our model has a strong generalization ability and can provide policymakers with accurate data support across regions in the future, which is important for scientific research and environmental policy formulation.
\subsection{Mapping the height of primary forest canopy in the fourth world-level giant tree distribution area}
\begin{figure*}[tp]
	\centering
	\includegraphics[width=1\textwidth]{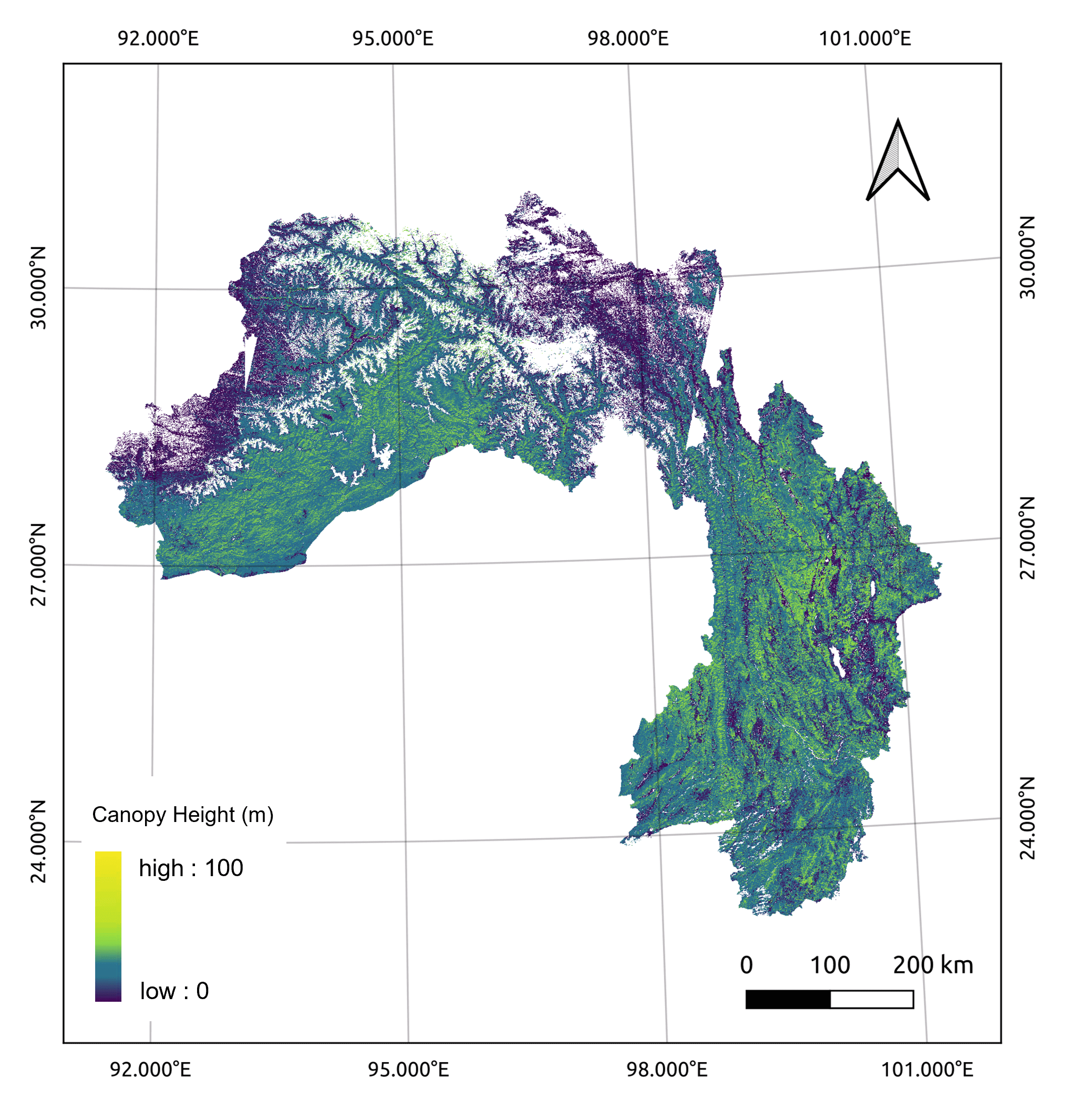}
	\caption{Primary forest height of the fourth world-level giant tree distribution area predicted by PRFXception}
	\label{Fig. 14}
\end{figure*}

As described in section 3.3, we trained the proposed PRFXception model and mapped the primary forest canopy height of the fourth world-level giant tree distribution center at 10m spatial resolution. Using Sentinel-2, ICESat-2, GEDI and other multi-source remote sensing data fusion as training data set, the modeling ability of PRFXception to plot the height of primary forest canopy on a regional scale is demonstrated. During training, the batch size is set to 5, the maximum epoch is set to 5000, and the learning rate is set to 0.0001. The loss function converges after about the 200th epoch. The training loss was slightly higher than the test loss, but the difference was less than 1, indicating that PRFXception did not overfit the training data to map the primary forest canopy height. Our proposed PRFXception method maps the height of the primary forest canopy in the fourth world-level giant tree distribution area (Fig. \ref{Fig. 14}).
\par We can see that the yellow area represents the higher canopy height, and the purple area identifies the lower canopy. The canopy height is relatively high (\textgreater 25m) in the southwest and central part of southeast Tibet and the central part of northwest Yunnan. On the whole, the canopy height in southeast Tibet gradually decreased from southwest to northeast and northwest, respectively. The higher canopy in the southwest direction is distributed in the range of more than 30m. The overall canopy height in northwest Yunnan is relatively low and is covered by forests below 30m. The vertical and fragmented distribution of the higher canopy may be related to the special geomorphology in Hengduan Mountain region. The climate in southeast Tibet and northwest Yunnan is complex and diverse, and the precipitation, temperature and sunshine conditions in different regions are quite different. In mountainous areas, the mountains are high and the terrain is complex. Different terrain conditions have different effects on the growth and development of forests. For example, in mountainous areas, forest growth is limited by topographic conditions, and the canopy height is generally low. In the valley area, the forest growth conditions are relatively good, and the canopy height is generally high.
\par
\begin{figure*}[tp]
	\centering
	\includegraphics[width=1\textwidth]{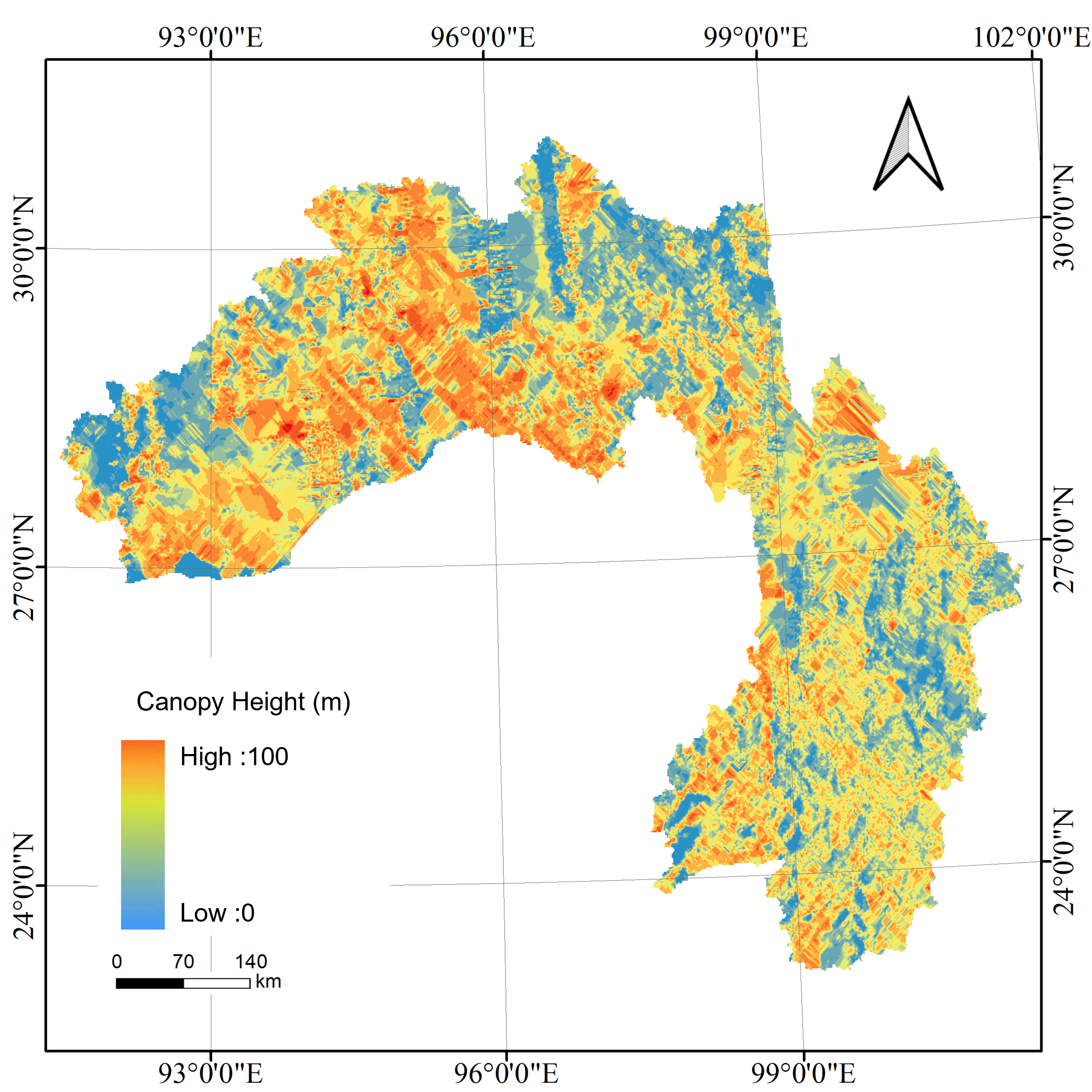}
	\caption{Primary forest height based on interpolated fourth world level giant tree center}
	\label{Fig. 15}
\end{figure*}

 Fig. \ref{Fig. 15} shows the primary forest canopy height of the fourth world level giant tree distribution area with GEDI and ICESat-2 footprint distribution plotted by our Kriging interpolation method. The spatial heterogeneity of satellite-borne LiDAR data may not be well dealt with by interpolation, especially when the data points are not evenly distributed or have nonlinear tendencies. This can cause the interpolation results to be discontinuous or inaccurate in some areas. The spatial resolution of the interpolated map is relatively low, which may lead to reduced accuracy in regression of canopy height and will further lead to increased uncertainty in canopy height prediction. Compared with Fig. \ref{Fig. 14}, it can be seen that our PRFXception model belongs to the extrapolation method, and the map of canopy height drawn has more abundant texture information, and the spatial change of height is smoother and more uniform. The method presented in this paper can explain large-scale geographical and ecological trends, such as forest, plateau, mountain, plain, mountain, snow mountain and other landforms and spatial distribution. Compared with the extrapolation method based on PRFXception, although the interpolation method can be used to plot the canopy height, it does not show the contour, texture and detail changes of the overall terrain well, and the results may be less reasonable and accurate.
\par
\begin{figure*}[tp]
	\centering
	\includegraphics[width=1\textwidth]{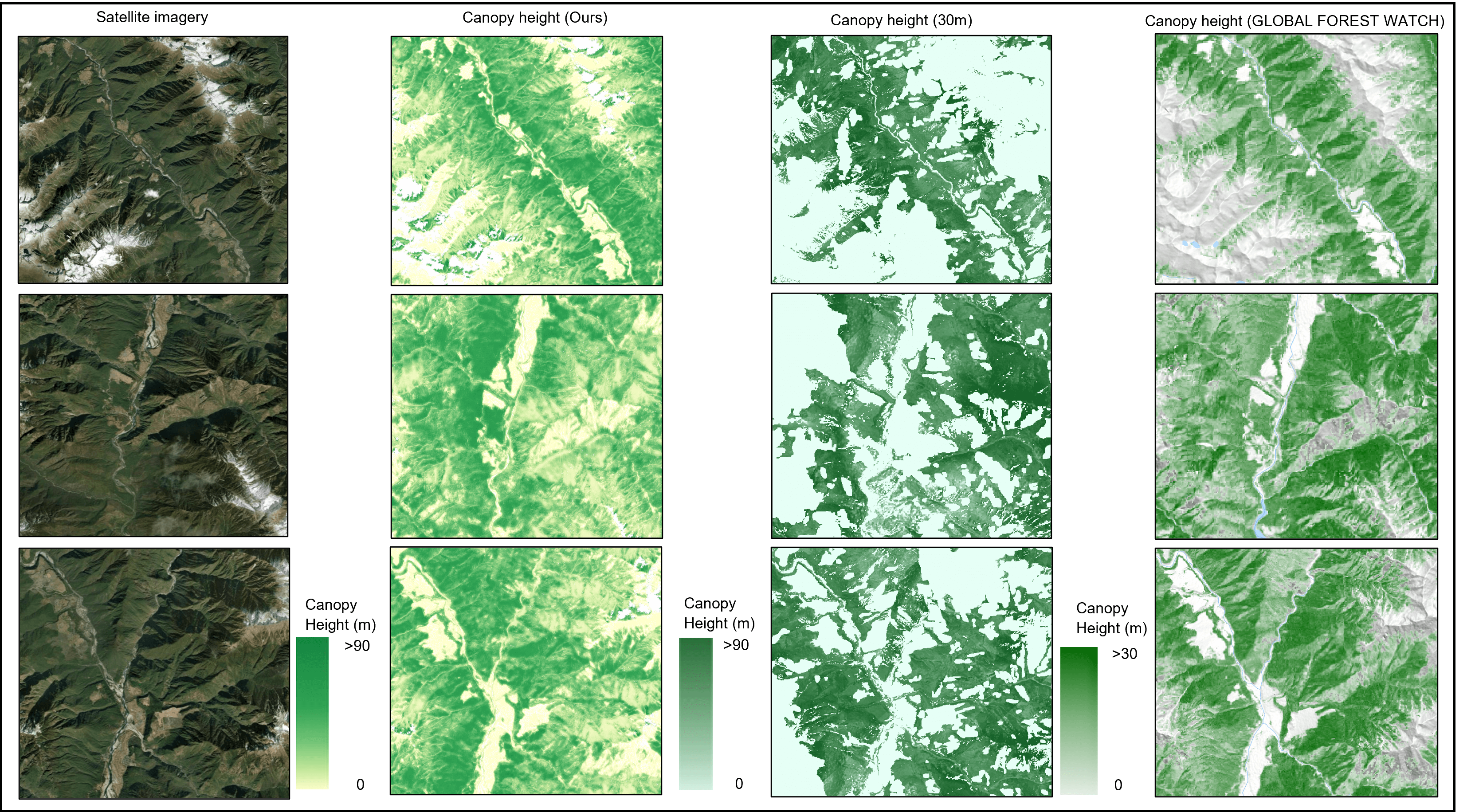}
	\caption{Compare canopy heights from different data sources (a) high-resolution satellite images (b) our canopy heights (c) canopy heights at 30m spatial resolution (d) canopy heights from global forest observers}
	\label{Fig. 16}
\end{figure*}

 Fig. \ref{Fig. 16} shows the comparison of canopy height maps from three different sources. The first column is a high-resolution satellite image that provides a true color view, the second column is a 10m GSD canopy height mapped by PRFXception, the third column is a 30m GSD canopy height \citep{liu2022neural}, and the fourth column is data provided by GLOBAL FOREST WATCH. From the comparison results, it can be seen that our canopy height regression method shows more detailed information, such as the texture characteristics of the terrain and the spatial distribution of the forest, which demonstrates the high resolution and high precision of the canopy height prediction capability of PRFXception. Using true-color images as a reference, we used the mapping results from PRFXception to show a smoother canopy height transition. Our altitude changes are more reasonable and match the characteristics of true color satellite imagery.
\par
\begin{figure*}[tp]
	\centering
	\includegraphics[width=1\textwidth]{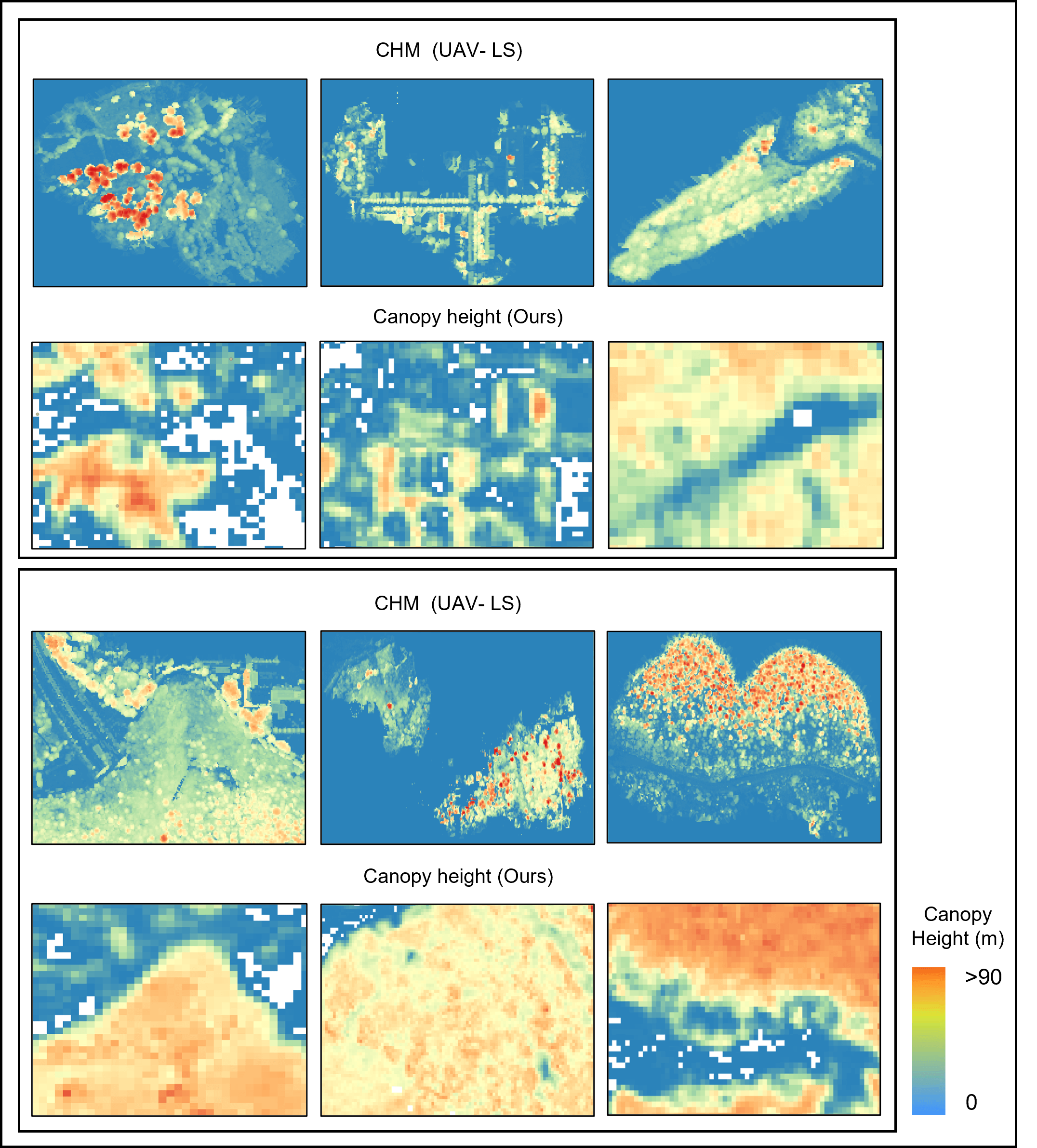}
	\caption{Forest canopy height correlation based on satellite observation data and UAV-LS point clouds}
	\label{Fig. 17}
\end{figure*}

 The first and third rows of Fig. \ref{Fig. 17} represent the canopy height as scanned by the drone, and the second and fourth rows are the canopy height predicted by PRFXception. The UAV-LS point clouds CHM has a spatial resolution of 1-2m, providing a very detailed distribution of canopy height, capable of showing small detail changes. The canopy height with 10m spatial resolution predicted by us from Sentinel-2 is consistent, similar and close to the UAV-based CHM, such as the change of the overall contour and the spatial distribution of height. This means that there is a high correlation between the predicted spatial distribution and the actual observation, indicating that PRFXception can accurately capture the main characteristics of the canopy height distribution of the primary forest and reliably infer the canopy height change. Due to the limitation of observation platform and spatial scale, the canopy height based on satellite observation data is not as high as that based on UAV-LS point clouds in terms of spatial resolution. Our method can still reflect the height structure and spatial distribution of forest canopy, which not only shows good consistency, but also is not limited by time and space.
\subsection{Potential application examples}
\subsubsection{Found potential world level giant tree individuals and communities}
World-class giant trees are the embodiment of the limits of individual growth. It is found that the search for world-level giant trees can better understand the growth potential of trees under specific environmental conditions, which is of great significance for ecological research. Giant tree individuals are usually older, and they play the role of "time witnesses" in the ecosystem, which is of unique value for the study of long-term environmental change and forest ecological succession. In forest communities, giant trees are essential for maintaining the structure and function of the forest. It plays a role in regulating the water cycle, providing biological genetic resources, maintaining soil stability, and regulating climate. Giant tree communities can provide a basis for ecological protection and help develop scientific and rational forest management strategies, especially in the face of forest degradation and ecological destruction. The protection of giant trees is not only beneficial for maintaining the current ecological balance, but also vital for combating climate change and protecting soil and water resources \citep{monastersky2023finding,ren2024conserving}. The study of individual and community distribution of giant trees is helpful for us to understand the mechanism of species coexistence in forest ecosystems, environmental factors of tree growth, and the impact of human activities on the natural environment. In the study of forest ecosystems and biodiversity, it is extremely important to understand and discover the individual and community distribution of world-level giant trees. As a key component of forest ecology, world level giant trees not only play a vital role in the maintenance of biodiversity, the stability of forest structure and function, and carbon cycling and storage, but also, they are living records of natural history and environmental meteorological changes.
\par
\begin{figure*}[tp]
	\centering
	\includegraphics[width=1\textwidth]{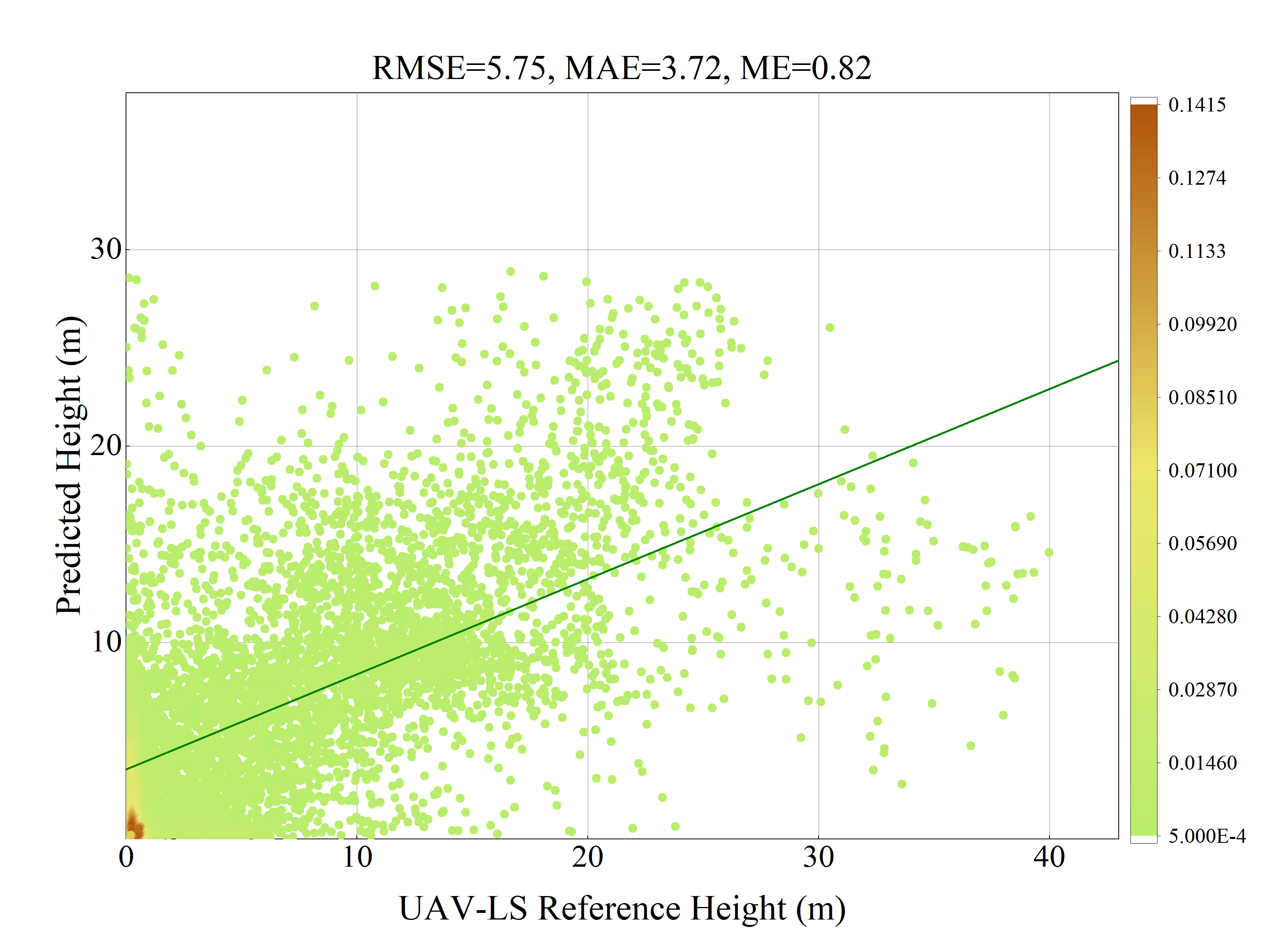}
	\caption{Using UAV-LS reference value model and point density map to predict canopy height}
	\label{Fig. 18}
\end{figure*}

 We use a custom depth-separable convolutional neural network (PRFXception) to predict undiscovered or potentially world-level giant tree individuals and communities in the study area. We will use the forest community point clouds scanned by UAV-LS as a new training data source to add to the training dataset of GEDI and ICESAT-2 in southeast Tibet. 465486 training samples were extracted from the CHM generated from the UAV-LS point clouds, and the predicted height of the corresponding PRFXception output was obtained. A total of 3281705 training samples were obtained in the experiment. Finally, training datasets including GEDI, ICESat-2, UAV-LS point clouds, and Sentinel-2 were produced to retrain PRFXception. We used the survey data of (GEDI, ICESat-2, UAV-LS), (GEDI, ICESat-2), (UAV-LS) and 227 permanent sample sites as reference values, respectively, to verify the accuracy of driving PRFXception to predict canopy height after integrating training data set into UAV-LS point clouds. Fig. \ref{Fig. 18} shows the relationship between the predicted height values and the reference values.
\par The following figure is the scatter density map obtained by using the UAV-LS reference value training model to predict the region, RMSE=5.75m, MAE=3.72m, ME=0.82m. Compared with GEDI \& ICESat-2 of "4.1" as a reference value, the reference height using UAV-LS has higher accuracy. From this scatter density plot, we can observe that most of the points are distributed near the diagonal, and as the reference height increases, the gap between the predicted value and the reference value gradually increases. Especially in the region of 30 \textasciitilde 40m, for the higher reference height, the model prediction height will be low and the accuracy will decline.
\par
\begin{figure*}[tp]
	\centering
	\includegraphics[width=1\textwidth]{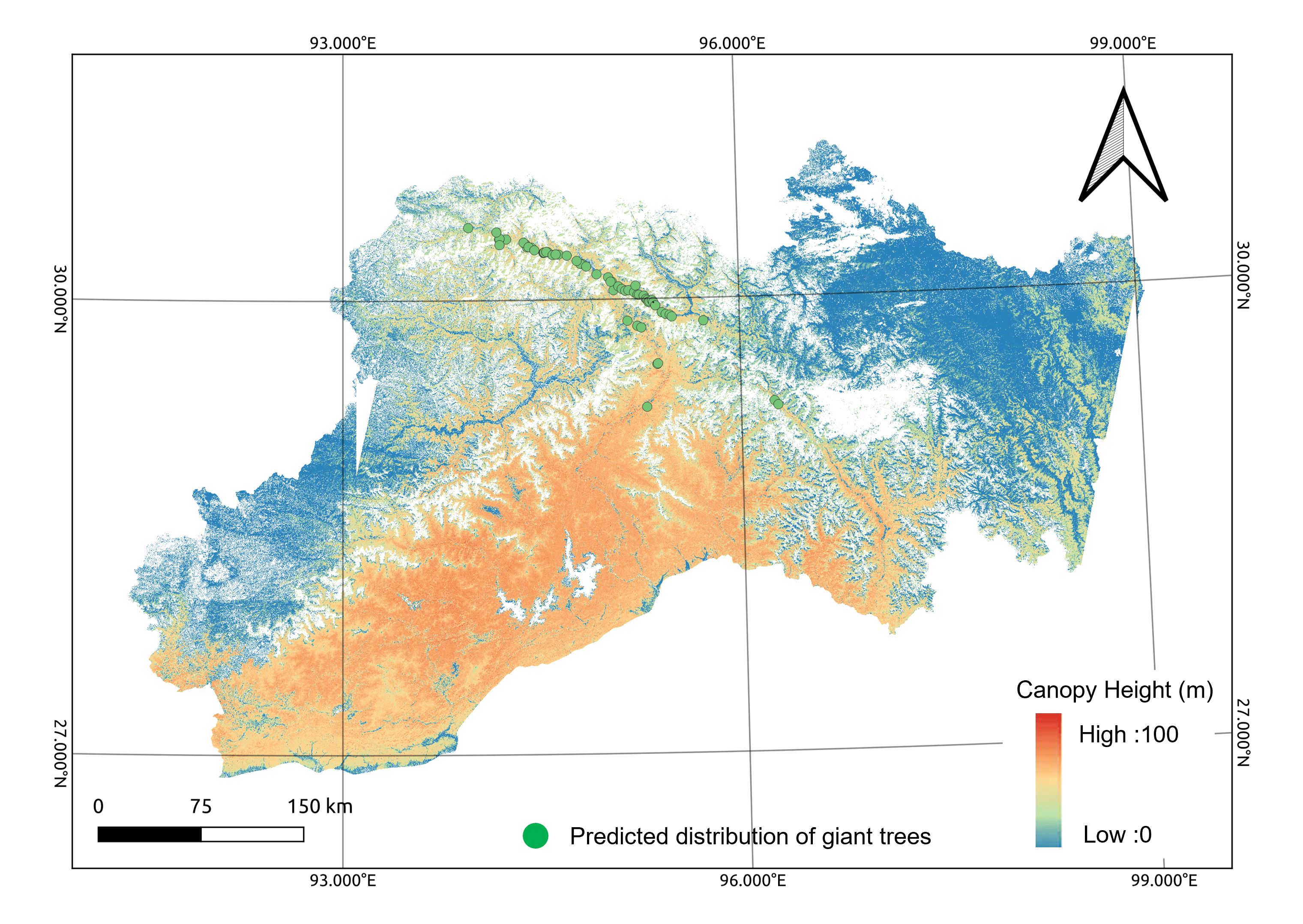}
	\caption{Map of potential world-level giant tree communities}
	\label{Fig. 19}
\end{figure*}

 Surprisingly, we found the geographic location of the tallest tree in Asia and its forest community in the canopy height map drawn by the UAV-LS point cloud-driven deep learning modeling on the basis of existing training data, and verified the geographical location of the tallest tree in Asia. And its forest community through our field investigation as a reference value. This result provides a valuable reference for us to discover more world level giant tree individuals and communities. On the basis of 10m resolution mapping of primary forest canopy height by PRFXception, we produced a vector map of potential distribution of giant trees in the fourth world-level giant tree distribution area (Fig. \ref{Fig. 19}). To find as many world-level giant tree individuals and communities as possible, the reference height value and predicted value were used for inter-zone statistics, and the accuracy of each interval was calculated as the possibility index. Potential giant trees are mainly distributed in river valleys, which may be related to soil fertility, water supply and climate in river valleys. These factors may have provided the right environmental conditions for the growth of the giant trees, resulting in a more concentrated and dense distribution in the valley. This concentrated distribution is important for understanding the formation and evolution of giant tree ecosystems and protecting these precious resources.
\par
\begin{figure*}[tp]
	\centering
	\includegraphics[width=1\textwidth]{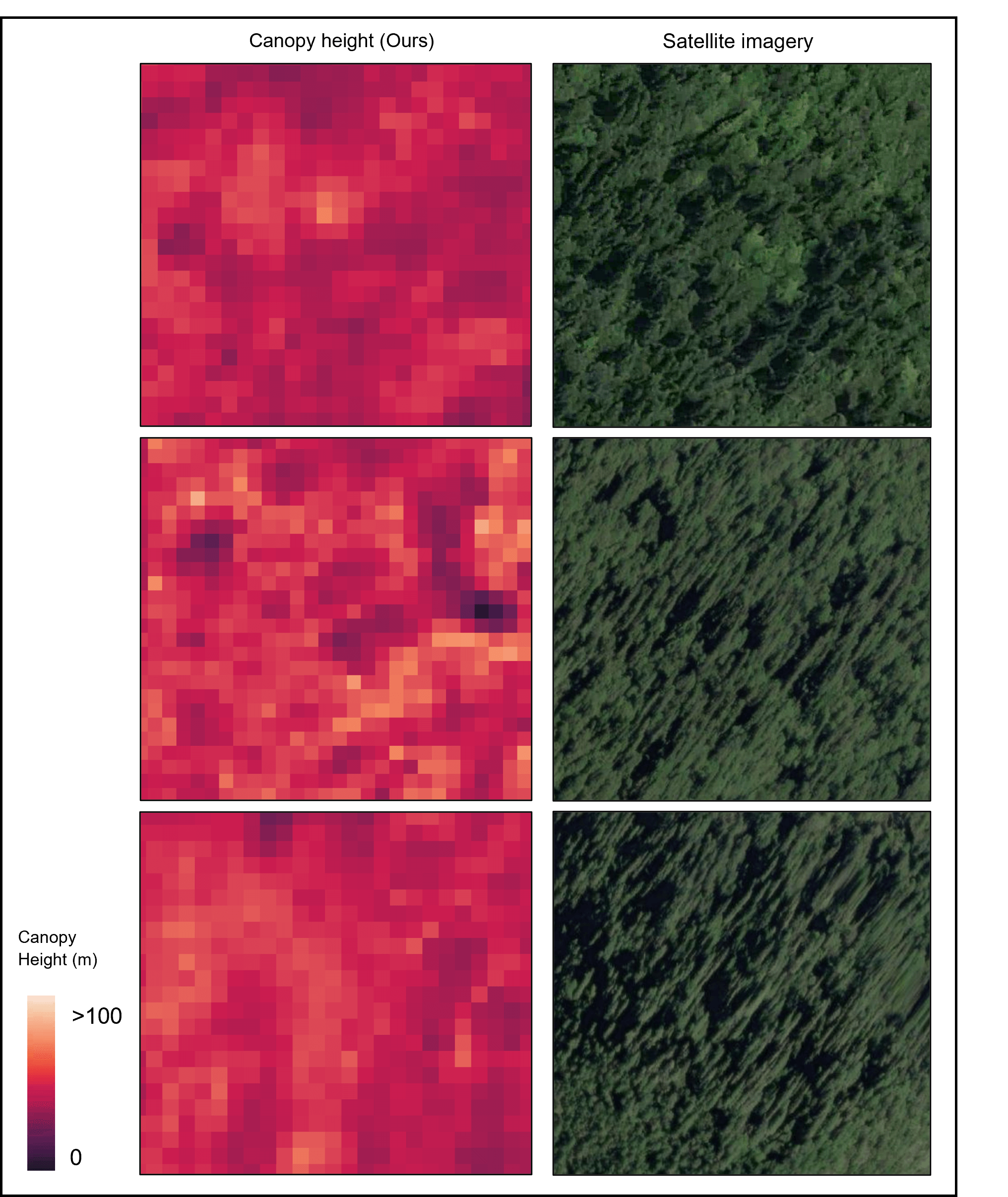}
	\caption{Map of canopy height at the site of giant tree community compared with satellite image}
	\label{Fig. 20}
\end{figure*}

 Surprisingly, the paper found two previously undiscovered communities of giant trees. Fig. \ref{Fig. 20} shows a raster map of the height of the giant tree community with the corresponding high-resolution satellite image. The first line is our field survey of the tallest trees in Asia. The second and third lines are the two giant tree communities that we newly discovered through this paper. From the satellite optical image of the forest community with the tallest trees in Asia, it can be observed that the trees in this region are tall and dense, and the spatial distribution pattern of canopy height in the grid map is reflected. The matching and similarity between the canopy height grid map of the community and the satellite image indicate the effectiveness of our method in discovering the tallest tree community, because the texture characteristics of the satellite image are consistent with the height grid characteristics. This significant matching and contrast effect proves the accuracy and reliability of the raster map, which means that our model has a good prediction effect.
\par The discovery of giant tree community is based not only on the height grid map of the forest community where the tallest trees in Asia are located and the texture, shape, tone, shadow, structure, contour and other features of the corresponding high-resolution satellite images, but also on the remote sensing images of drones and field investigation data as the basis for the judgment of giant tree community. Therefore, it is reasonable to assume that the second and third lines (Fig. \ref{Fig. 20}) show two previously undiscovered communities of giant trees. Based on the number and distribution of canopy height image elements, height spatial distribution characteristics and corresponding image texture characteristics of the tallest tree communities in Asia, we concluded that the probability of the canopy height of the two giant tree communities is 89\% in the range of 80-100m, and there is a certain probability of the growth of giant tree individuals higher than the tallest tree communities in Asia.
\par Our judgment was also combined with environmental factors such as temperature, precipitation, climatology, elevation, and soil as shown in the second and third lines (Fig. \ref{Fig. 20}). The two giant tree communities found in this paper are located in the hinterland of the mountain range, near the tributaries of the Brahmaputra River, with sufficient water resources and fertile soil. Nature reserves have been established in these areas, and human activities have less intervention, which is conducive to the growth of vegetation. The region’s ecology is relatively well preserved, with unique biodiversity and an understudied resource of giant trees. According to the distribution of giant tree communities, it is mainly distributed in the YTGC. There may be more world-level giant tree individuals and communities. The stable environment that may exist in the region reduces the risk of large trees suffering from catastrophic events, such as floods, landslides or extreme weather events. Under the barrier of the mountains on both sides of the YTGC, the warm and wet air of the Indian Ocean can enter the plateau to the north, bringing abundant precipitation and heat to the YTGC area. According to the data released by the National Meteorological Information Center (https://data.cma.cn/), the average monthly wind speed at this location is no more than 2 m/s. Under the condition of low wind speed, trees can develop strong and stable structures, promote the growth of roots and thus improve the stability of trees. Trees growing in moderate winds may show better adaptability and resistance to stress. The monthly precipitation exceeds 120 mm, the annual average temperature is 8-13 ℃, the annual relative humidity (60\%-80\%) is relatively higher than that of the surrounding area (40\%-50\%), and the annual sunshine duration (1500-2500 hours) is relatively lower than that of other surrounding areas (3000-4000 hours). The warm and humid environment helps to promote cell division and elongation, which directly affects the growth rate of trees, and these meteorological factors provide favorable conditions for the growth of world-level giant trees and communities. The soil formed by sedimentation in the YTGC basin is usually very fertile and rich in organic matter and mineral nutrients, providing the nutrients needed for the giant trees. We passed the National Soil Database (http://www.soilinfo.cn/map/) to see the area the NPK element content of soil, organic matter content, the net primary productivity indicators than other areas, such as the relatively remote geographical position, means that these areas are relatively not influenced by the excessive human activities, It provides good natural growth conditions for the growth of world-level giant trees and communities.
\subsubsection{Forest biomass and carbon stock}
Our method has the potential to predict biomass and carbon stocks within the world's largest tree distributions, providing valuable data support for ecological conservation, natural resource management and climate change research. As the impact of global climate change on ecosystems intensifies, the conservation and research of these giant trees is integral to developing strategies to combat it and promote sustainable development. In the context of GEDI, ICESat-2, Sentinel-2 and other Earth observation mission objectives, our proposed method can be used to fill the gap in forest canopy height data products in world-level giant tree areas, especially in areas not covered by GEDI and ICESat-2 orbits. The new high-resolution canopy height dataset in this paper can help advance at least two major downstream applications at a regional scale, namely biomass and carbon stock modeling. Canopy height is a key indicator of regional or global above-ground carbon stock in the form of biomass. At a regional scale, we compare the canopy height map with dense above-ground carbon density data, which was generated by ESA (Fig. \ref{Fig. 20}). The biomes within the distribution of the world's giant trees are undergoing unprecedented changes related to human activities, exacerbated by climate change. Understanding the carbon content of these forests and how that carbon has changed - and is changing - is an important step in securing their long-term future and tackling climate change. The key advantage of our method to map the height of primary forest canopy from satellite observations is that it provides a regionally consistent approach with repeated and consistent measurements from space. While the definition of biomass depends not only on canopy height, but also on vegetation parameters such as DBH and wood density, and these derivations require more field data and threshold calibration, a series of works will help track changes in biomass distribution and density over time, which in turn will inform policies that promote carbon reduction and forest conservation initiatives. For example, the United Nations Plan to reduce emissions from deforestation and degradation \citep{crockett2023structural,tang2021spatiotemporal}. We observed that canopy height estimates for Sentinel-2 can even predict carbon density in old-growth forest areas, with canopy heights up to 65m. Notably, the relationship between carbon and canopy height is sensitive up to 60m. We note that while it is technically possible to map biomass at a sampling distance of 10 meters above ground, this may not be practical.
\begin{figure*}[tp]
	\centering
	\includegraphics[width=1\textwidth]{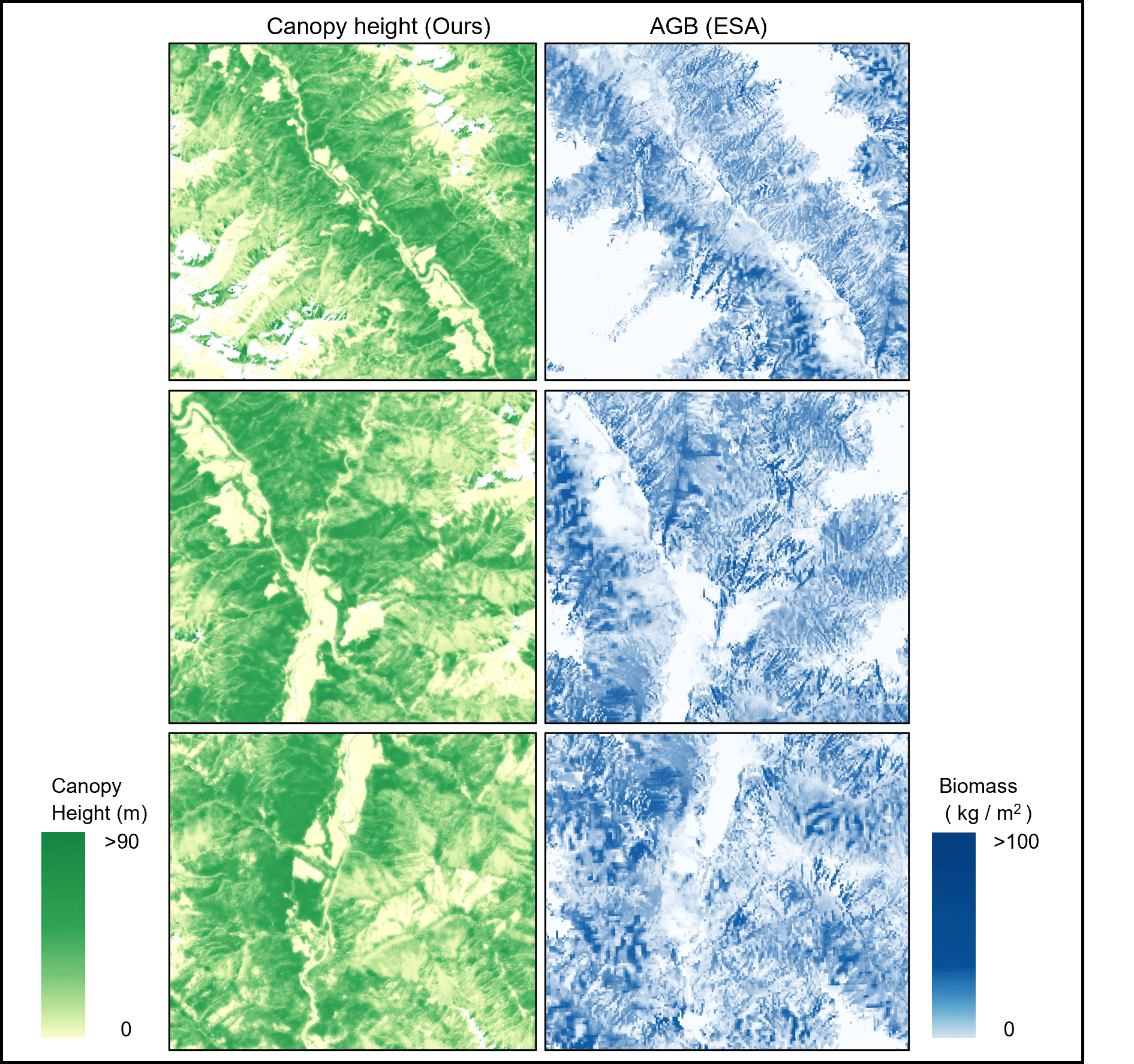}
	\caption{Comparison of our predicted canopy height map with the ESA biomass map}
	\label{Fig. 21}
\end{figure*}
\par Our proposed method has the potential to accurately predict biomass and carbon stocks in primary forest areas, providing important data support for ecological conservation, resource management and climate change research. Our study fills a gap in the world-level giant tree distribution for Earth observation missions, providing high-resolution canopy height datasets that can be used for biomass and carbon stock modeling. Canopy height was compared with above-ground carbon density data to understand the global distribution of above-ground carbon stocks, particularly in uncovered areas. This paper will provide regionally consistent measures of biomass change for satellite observations, and while biomass derivation requires more field data and threshold calibration, our work will help track biomass change over time to inform carbon reduction and forest conservation policies.
\par Fig. \ref{Fig. 21} shows the comparison between forest canopy height and forest above-ground biomass. When comparing forest canopy height and biomass, we can observe that there is a significant correlation between them. The forest canopy height map shows the vertical height distribution of trees in the forest, while the biomass map shows the total biomass of vegetation in the same area. Although the information expressed by the two is different, when we observe their texture, we will find that the two have a high similarity in the fluctuation, density and spatial distribution displayed on the image. The similarity of textures suggests that there may be a deeper correlation between forest canopy height maps and biomass maps. This association may arise from the structure and organization within the forest, such as the growth of trees, the extent of canopy cover, and population density. When forest canopy height increases, it generally implies a corresponding increase in population density and biomass within the forest, possibly due to more complex forest structures, an observation that provides insight into the complex relationship between forest structure and function. A thorough understanding of the relationship between forest canopy height and biomass is essential for ecosystem management and conservation, and can help us better understand the dynamic changes of forest ecosystems and develop effective conservation and management strategies \citep{oehmcke2024deep,pelletier2024inter}.
\par Compared with 30m spatial resolution canopy height, the 10m resolution map drawn by our method has obvious advantages \citep{liu2022neural,scheeres2023distinguishing}. Higher resolution brings finer texture structure, and detailed information is conducive to finding the existence of higher world-level giant trees. Our method is particularly suitable for areas with large rolling terrain and dense vegetation, such as subtropical or tropical regions, where a single canopy may exceed 10 m pixels. It is recommended to model biomass at a coarser spatial resolution (e.g. 0.25 ha) suitable for capturing changes in densely vegetated areas. However, high-resolution canopy height data have great potential to improve regional or global biomass estimates by providing descriptive statistics on vegetation structure within local areas. Our model can be deployed in the future with high temporal resolution (annual or quarterly) to map changes in the height of primary forest canopy over time, for example, changes in acquired carbon stocks and to estimate carbon emissions from global land use change, currently primarily deforestation.
\subsubsection{The integrity of biodiversity}
The integrity of biodiversity can reflect the quality of the habitats of animals and plants in nature reserves, and the height of forest canopy is closely related to the calculation of ecosystem biodiversity integrity. Higher canopies tend to be associated with richer ecosystems, higher species diversity, and more complex biomes, and are associated with the height of tree species in the forest \citep{silveira2021spatio,wang2019remote}. The higher the biodiversity integrity of the area with crown height, the lower the biodiversity integrity. Because large trees provide rich living space and resources for various plants and animals, they promote the formation and maintenance of diverse biological communities.
\par
\begin{figure*}[tp]
	\centering
	\includegraphics[width=1\textwidth]{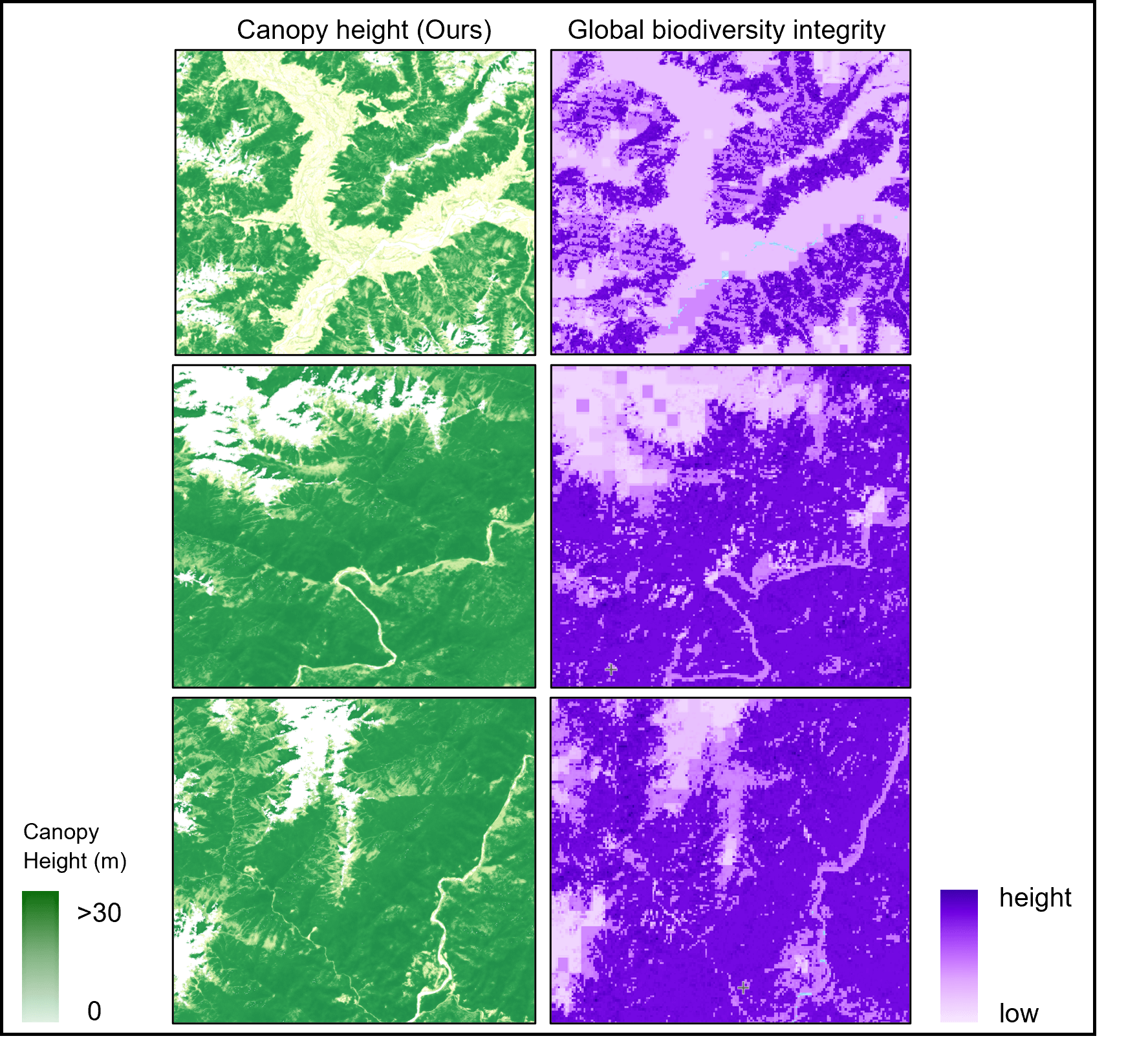}
	\caption{Our canopy height versus biodiversity integrity}
	\label{Fig. 22}
\end{figure*}

 Fig. \ref{Fig. 22} shows a comparison between the canopy height we plotted (left) and the biodiversity integrity from Global Forest Watch (right). It can be seen that both show similar spatial distribution patterns and positive correlations, indicating that our method can reflect biodiversity integrity to some extent. Higher canopy is associated with higher biodiversity integrity. Because more mature forests typically provide more habitat types and a richer variety of organisms. Meanwhile, the river. Topographic features such as roads are also clearly visible in the two figures and produce similar patterns in corresponding positions, which also confirms the spatial consistency between the two. In addition, our method can also be used as one of the important indicators to assess the health of primary forest ecosystems, because higher trees tend to mean that the ecosystem is less stressed and disturbed, expressing the integrity and authenticity of the primary forest. Monitoring forest canopy height using our method can therefore yield important insights on biodiversity, ecosystem health and environmental change, providing a scientific basis for ecological conservation and sustainable management.
\section{Conclusions}
This study is the first endeavor to map the height of the primary forest canopy within the distribution area of the fourth-world giant trees, where Asia’s tallest trees flourish. In this paper, we have presented a data-driven deep learning based method that fuses both satellite-borne LiDAR and satellite images (GEDI, ICESat-2 and Sentinel-2) to delineate the canopy height of primary forest. Specifically, we propose a depth-separable convolutional neural network, PRFXception, augmented with pyramid-sensitive field, tailored for this task. PRFXception utilizes Sentinel-2 optical images to estimate canopy height in the distribution area of giant trees, achieving a remarkable spatial resolution of 10 meters. We integrate data from GEDI, ICESat-2, Sentinel-2, UAV-LS, and field surveys for the first time, facilitating comprehensive mapping of canopy height of primary forest. Our fusion of spaceborne LiDAR and satellite image underpins the PRFXception model, facilitating validation and calibration against ground sample surveys and UAV-LS point clouds across the giant tree distribution area. Our experiments have demonstrated that the proposed approach holds promise for identifying world-class giant tree specimen and communities. Notably, our analysis uncovers two previously unknown world-level giant tree communities, with an 89\% probability of falling within 80-100m height range, suggesting the potential for individuals surpassing Asia’s tallest trees. 
\par While previous studies have primarily focused on extending the regional-scale canopy height models to a global scale, our emphasis lies in developing a high-performance regional-scale primary forest canopy height model. In practical applications, our model can be used to monitor forest height without reliance on ground data, mitigating the inherent underestimation bias associated with high canopy. Theoretically, our model can be refined without necessitating local calibration, making it particularly suitable for tropical or subtropical mixed primary forest areas harboring world-level giant trees. This study provides compelling scientific validation for designating Southeast Tibet-Northwest Yunnan as the fourth world-level giant tree distribution hub. Such recognition is pivotal in supporting global climate and sustainable development initiatives and promoting the inclusion of the Brahmaputra Grand Canyon National Nature Reserve within the national park protection framework of China.

\section*{Acknowledgement}
This article thanks Professor Qinghua Guo of Peking University and his team members for their contributions to the discovery of the tallest tree in Asia. We would like to thank the Chinese Giant Tree Research Team, which consists of Institute of Botany, the Chinese Academy of Sciences, Chenshan Center-Shanghai Chenshan Botanical Garden, China Environmental Protection Foundation, "Wild China" Studio, and Forestry and Grassland Bureau of Zayü County, Nyingchi City, Tibet Autonomous Region, for their contributions to the Chinese Giant Tree Research. This work was supported by Tibet Autonomous Region Science and Technology Plan Project (XZ202301YD0043C); Key Project of National Natural Science Foundation of China (4233000283); National Postdoctoral Innovative Talent Support Program (BX20220038).
\section*{Declaration of Competing Interest}
The authors declare no conflict of interest.
\section*{Data availability}
Data will be made available on request.

	
\bibliographystyle{elsarticle-harv}

	\bibliography{mmm.bib}
	
	
	
\end{document}